\documentclass{article}
\usepackage{stfloats}
\usepackage{tabularx} 
\usepackage{PRIMEarxiv}
\usepackage{amsmath}
\usepackage{amssymb}
\usepackage{booktabs} 
\usepackage{multirow}
\usepackage[utf8]{inputenc} 
\usepackage[T1]{fontenc}    
\usepackage{hyperref}       
\usepackage{url}            
\usepackage{booktabs}       
\usepackage{amsfonts}       
\usepackage{nicefrac}       
\usepackage{microtype}      
\usepackage{lipsum}
\usepackage{fancyhdr}       
\usepackage{graphicx}       
\graphicspath{{media/}}     

\title{A Real-time Concrete Crack Detection and Segmentation Model Based on YOLOv11}

\author{
  Shaoze Huang* \\
  MSE, AHUT, China\\
  szhuang0601@163.com\\
   \And
  Qi Liu \\
  IoT, AHUT, China\\
  liuqi04@126.com\\
     \And
  Chao Chen \\
  IST, SDUST, China\\
  cchen1107@126.com\\
     \And
  Yuhang Chen \\
  MSE, AHUT, China\\
  yhchen2508@163.com\\
}

\begin{document}
\maketitle

\begin{abstract}
Accelerated aging of transportation infrastructure in the rapidly developing Yangtze River Delta region necessitates efficient concrete crack detection, as crack deterioration critically compromises structural integrity and regional economic growth. To overcome the limitations of inefficient manual inspection and the suboptimal performance of existing deep learning models, particularly for small-target crack detection within complex backgrounds, this paper proposes YOLOv11-KW-TA-FP, a multi-task concrete crack detection and segmentation model based on the YOLOv11n architecture. The proposed model integrates a three-stage optimization framework: (1) Embedding dynamic KernelWarehouse convolution (KWConv) within the backbone network to enhance feature representation through a dynamic kernel sharing mechanism; (2) Incorporating a triple attention mechanism (TA) into the feature pyramid to strengthen channel-spatial interaction modeling; and (3) Designing an FP-IoU loss function to facilitate adaptive bounding box regression penalization. Experimental validation demonstrates that the enhanced model achieves significant performance improvements over the baseline, attaining 91.3\% precision, 76.6\% recall, and 86.4\% mAP@50. Ablation studies confirm the synergistic efficacy of the proposed modules. Furthermore, robustness tests indicate stable performance under conditions of data scarcity and noise interference. This research delivers an efficient computer vision solution for automated infrastructure inspection, exhibiting substantial practical engineering value.
\end{abstract}


\section{Introduction}
The Yangtze River Delta, situated at the estuary of the Yangtze River, encompasses Shanghai, Jiangsu Province, Zhejiang Province, and Anhui Province. As one of the most economically vibrant and dynamic regions in China, it plays a pivotal role in the country's economic development and is a crucial area for research and innovation. In the early days, the transportation infrastructure in this area mainly consisted of stone bridges and dirt roads. Later, with the development of industry, railway bridges and highway bridges were gradually built to meet the transportation needs. Rapid urbanization in the Yangtze River Delta has accelerated infrastructure aging. Critical structures like bridges and tunnels now face deterioration issues while supporting transportation demands. Concrete crack issues, as a common type of damage, seriously affect the structural safety and durability. Therefore, timely and accurate detection of cracks has become an important task for infrastructure maintenance and safety management\cite{1,2}.

Traditional concrete crack detection mainly relies on visual inspection by humans. However, this method is time-consuming, labor-intensive, and easily affected by human factors, leading to subjectivity and inconsistency in the detection results\cite{3,4}. Moreover, it is difficult to inspect special locations such as bridge piers and the interiors of tunnels, which increases safety hazards\cite{5}.

\begin{figure*}[t!]  
  \centering
  \includegraphics[width=16cm]{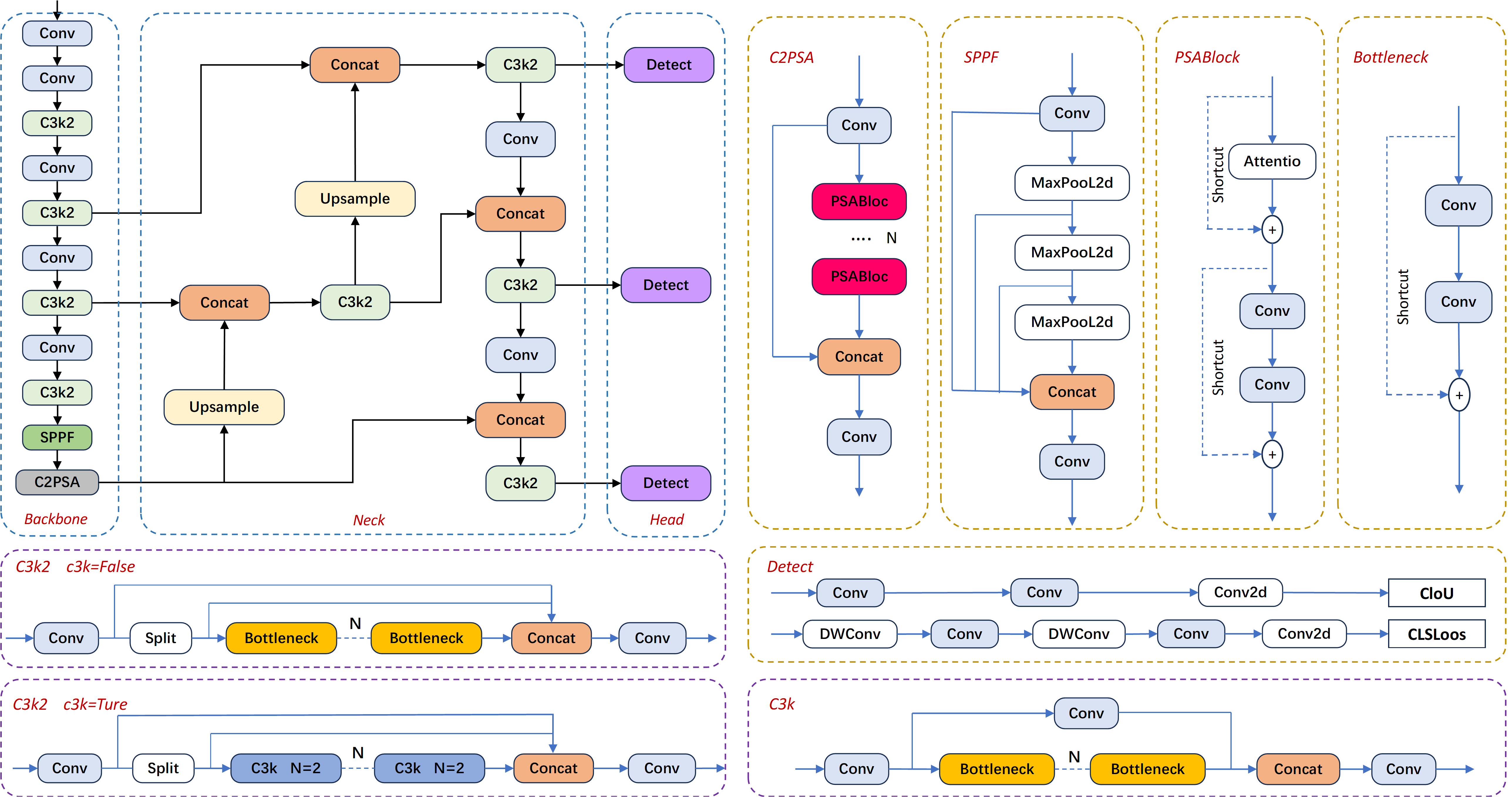}  
  \caption{YOLOv11 Model Architecture.\label{fig1}}
\end{figure*}

In recent years, deep learning has achieved remarkable advancements in the field of computer vision, with object detection being a prime example. YOLO, serving as an efficient real-time object detection framework, has found extensive applications in traffic monitoring, security surveillance, and industrial detection\cite{6,7}. As the latest iteration, YOLOv11 has emerged as a preferred solution for concrete crack detection, thanks to its efficient feature extraction capability and superior real-time processing performance\cite{8,9}.Its model structure is shown in Figure \ref{fig1}.

This research, leveraging the YOLOv11 model, has developed an intelligent system dedicated to crack detection in the infrastructure within the Yangtze River Delta region. The system is capable of swiftly and precisely identifying concrete cracks in bridges, tunnels, and water conservancy facilities, offering a more efficient and accurate approach in comparison with traditional methods. The datasets utilized encompass a diverse range of crack images from the Yangtze River Delta, guaranteeing the model's generalization ability and precision\cite{10,11}. This study holds substantial importance for enhancing detection efficiency and offers theoretical and technical backing for intelligent infrastructure maintenance\cite{4,12}. Through this research, it is intended to present an innovative solution for the safety management of infrastructure not only in the Yangtze River Delta but also in broader regions, and to advance intelligent detection and maintenance\cite{13,14}. Against the backdrop of rapid urbanization, expansive transportation networks, and the limitations of visual inspection, this study strives to tackle the challenge of concrete crack detection in complex backgrounds by creating an efficient deep convolutional neural network (CNN) model. The key contributions of this paper are as follows:

\begin{itemize}
\item \textbf{Backbone network}: Replace the original convolution with KernelWarehouse (KW) convolution, dynamically adjust the convolution kernel weights, enhance feature representation, reduce computational complexity, and improve the model's robustness as well as its ability to capture fine-grained features and perform multiscale fusion;
\item \textbf{Feature pyramid network}: Integrate a triple attention (TA) mechanism to enhance feature extraction and optimize multiscale object detection, thereby improving the detection capabilities for small and occluded objects; strengthen pixel-level feature representation, enhance the utilization efficiency of contextual information, and optimize the segmentation performance for small objects and complex backgrounds, thus improving the overall model performance and robustness;
\item \textbf{Loss function}: Construct an FP-IoU loss function for the object detection task. By employing adaptive penalty factors and interval mapping, it improves the precision of bounding box regression and pixel-level localization, enhances the detection and segmentation capabilities for small objects and low-quality samples, accelerates model convergence, increases training efficiency, and strengthens generalization and robustness.
\end{itemize}

\section{Related Work}

With the advancement of computer science, a substantial number of computer vision algorithms have been proposed and applied across multiple domains. Current crack detection models can be categorized into two main approaches: classical-based methods and deep learning-based methods\cite{15},\cite{47}.

\subsection{Traditional Crack Detection Methods}

In the field of crack detection, traditional image processing methods have formed a complete technical system. In early research, algorithms based on edge detection were widely explored. Dorafshan et al. compared the performance of spatial domain gradient detection and mathematical morphology gradient detection in concrete crack identification, and the results showed that the latter had better response efficiency under complex background conditions\cite{16}. To address the limitations of the Sobel operator, Zhang et al. increased six directional templates and improved the threshold segmentation algorithm, effectively improving the positioning accuracy and continuity of crack edges\cite{17}. The combination of morphological operations and threshold segmentation has also been proven to be effective. Jia's research on traditional methods showed that adaptive local thresholding combined with dilation and erosion operations could effectively suppress light interference and enhance crack features\cite{19}. Xu et al. used iterative threshold segmentation combined with morphological dilation to connect cracks, successfully eliminating the interference of pseudo-edges\cite{18}. Otsu's method has been widely used in threshold selection; for example, Fan optimized edge detection in noisy environments by combining Gaussian filtering with Otsu's method\cite{20}. In terms of algorithm performance evaluation, Li et al. used support vector machines (SVM) to classify crack shapes, combined with adaptive gray-scale transformation and geometric feature extraction, significantly improving the accuracy of distinguishing diagonal and mesh cracks\cite{21}. The inherent limitations of traditional methods have led to the emergence of multi-technology integration strategies. For example, Shi et al. combined pulsed eddy currents with electromagnetic acoustic transducers to achieve quantitative detection of cracks in metal components, effectively overcoming the limitations of single eddy current detection, which is susceptible to lift-off effects\cite{22}.

Although traditional crack detection methods play a fundamental role, they have inherent limitations: they heavily rely on manual feature engineering (such as hand-designed filters and thresholds), making it difficult to adapt to diverse crack shapes; they lack the ability to model complex contextual interactions, leading to false detections in noisy environments such as rebar shadows or surface stains; and they have poor generalization to different infrastructure scenarios, especially for low-contrast or blurry cracks. To overcome these challenges, the proposed YOLOv11-KW-TA-FP model introduces three synergistic innovations: KernelWarehouse convolution (KWConv) dynamically adjusts convolutional kernel weights at the unit level, autonomously capturing multi-scale crack features without human intervention; the triple attention mechanism (TA) coordinates spatial, channel, and long-range dependencies to suppress background interference while enhancing key crack features; and the FP-IoU loss function, combined with adaptive geometric penalties and non-monotonic attention mechanisms, improves the localization accuracy of low-quality samples. These innovations collectively break through the limitations of traditional methods, providing a robust automated detection solution for modern infrastructure.

\subsection{YOLO for Crack Detection}

In recent years, deep learning models have made breakthrough progress in the field of computer vision. Although the concept of deep learning was proposed long ago, its widespread application began with the introduction of graphics processing units (GPUs) in 2009\cite{23}. The remarkable parallel computing capabilities of GPUs have significantly accelerated the training process of deep neural networks in image classification and recognition tasks\cite{24}. As infrastructure continues to age, crack detection has become increasingly crucial for ensuring the safety of structures such as roads and buildings. Benefiting from its efficient and real-time detection characteristics, the YOLO series of models have been widely adopted in the field of crack detection.

The YOLO-DL framework\cite{25} integrates the YOLO and DeepLabv3+ network architectures, proposing a model suitable for multi-task recognition of concrete cracks. By introducing attention mechanisms and feature calibration modules, the model significantly improves its performance in crack classification, localization, and segmentation tasks, especially showing advantages in real-time detection. Hu et al. proposed a wind turbine damage detection method based on YOLO, which combines deep learning with UAV inspection technology to achieve high-precision, real-time damage detection and segmentation\cite{26}. In the field of concrete crack detection, compared with other mainstream algorithms such as Faster R-CNN and SSD, YOLO demonstrates stronger robustness and accuracy, especially in dealing with tiny cracks and imbalanced data distributions, where its performance is more stable\cite{27}.

To optimize the performance of YOLO in road defect detection, researchers have proposed a series of improved algorithms. For example, the improved method that combines a partial Transformer structure with multi-focus attention mechanism effectively improves detection accuracy while ensuring real-time performance\cite{28}. In the scenario of rock crack detection, the YOLO-based method successfully meets the challenges of complex underground environments and achieves accurate identification of crack types and features\cite{29}. DAPONet proposes an improved YOLO model that integrates a dual-attention mechanism for real-time road damage detection and achieves excellent performance on multiple public datasets\cite{30}. The CL-PSDD model, which utilizes contrastive learning techniques and the YOLO framework, demonstrates strong crack detection capabilities under the unsupervised learning paradigm\cite{31}.

However, challenges persist. Current models exhibit reduced precision in high-noise environments (e.g., construction sites with debris interference) and struggle with low-contrast micro-cracks. The requirement for extensive labeled datasets increases implementation costs, and further optimization is needed to balance precision with computational efficiency in resource-constrained field deployments.

\section{Proposed Method}

\begin{figure*}[t!]  
  \centering
  \includegraphics[width=\textwidth]{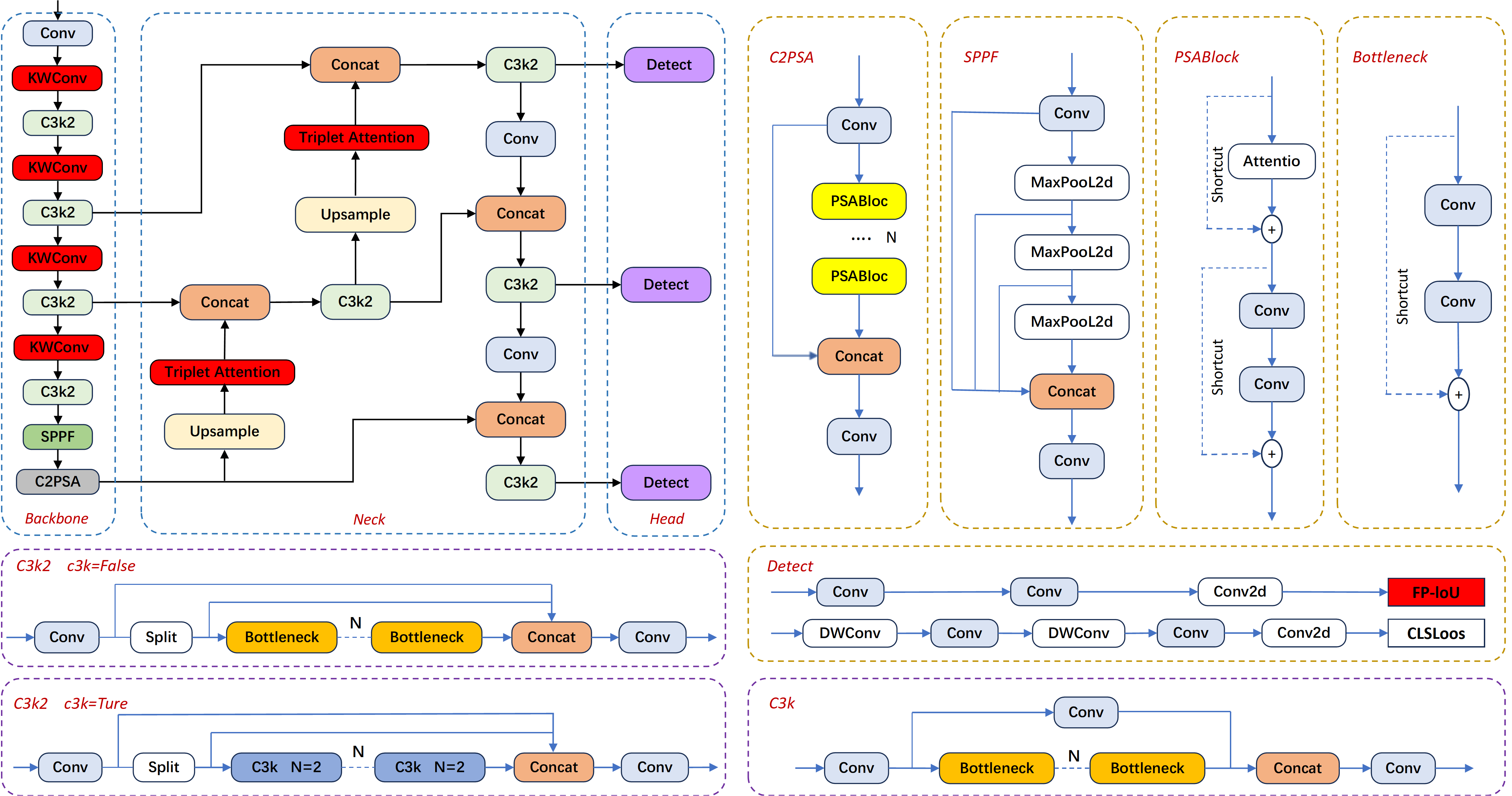}  
  \caption{YOLOv11-KW-TA-FP Model Architecture.\label{fig2}}
\end{figure*}

YOLOv11n, by streamlining network layers and reducing parameters and computations, delivers faster inference and a more compact model, fitting for resource - limited devices\cite{32,33}. Therefore, this paper selects it as the baseline model. Despite its excellent detection performance, YOLOv11n still has room for improvement in the concrete crack detection scenario. Based on YOLOv11n, this paper proposes the YOLOv11-KW-TA-FP model for concrete crack detection, the structure of which is shown in Figure \ref{fig2}. The red modules mark the key improvements to YOLOv11 in this study: swapping standard convolution in the backbone network with KernelWarehouse Convolution (KWConv), adding a TA mechanism module after each upsample module in the feature pyramid network (Neck), and substituting the original CIoU loss function with the FP-IoU loss function in the detection head (Head).

\subsection{KernelWarehouse Dynamic Convolution}

In object detection tasks, background interference remains a persistent challenge. To enhance the model's capacity to handle complex scenes and improve detection precision, this study integrates KernelWarehouse\cite{34} into the backbone network. Developed by Intel in August 2023, KernelWarehouse represents a generalized dynamic convolution framework that achieves an optimal equilibrium between parameter efficiency and representational power. As illustrated in Figure \ref{fig3}, the method comprises three core components: (1) adaptive kernel partitioning, (2) cross-layer warehouse sharing, and (3) the normalized attention function (NAF) for dynamic kernel fusion.

\subsubsection{Kernel partition}

Traditional dynamic convolution directly mixes complete convolution kernels, resulting in a linear increase in parameters with the number of kernels. KernelWarehouse innovatively splits a single convolution kernel evenly along the channel dimension into m non-overlapping kernel units, with each unit having a dimension of only 1/m of the original kernel. For example, when m=16, the number of parameters for each kernel unit is reduced to 1/16 of the original kernel. By refining the mixing granularity from the complete kernel to the kernel unit level, this method allows a significant increase in the number n of kernel units stored in the warehouse under the same parameter budget (e.g., n=108), breaking through the traditional method's limit of n<10. Each kernel unit is dynamically generated through linear combination, significantly improving parameter efficiency and feature representation capability.

\subsubsection{Warehouse sharing}

To enhance cross-layer parameter reuse, KernelWarehouse designs a hierarchical sharing mechanism: convolution layers in the same stage share the same warehouse E, while different stages use independent warehouses. In modern convolutional networks (such as ResNet and MobileNet), layers in the same stage usually have the same kernel size. The dimension of the unified kernel unit is determined by calculating the greatest common divisor of the kernel sizes of each layer. For example, the kernels of adjacent layers with sizes 3×3×64×128 and 3×3×128×256 can be uniformly split into kernel units of size 1×1×64×64. This strategy explicitly enhances the inter-layer parameter dependency, allowing the model to keep the overall parameter increase controllable when the total number of kernel units n increases. Experiments show that the wider the sharing range (cross-layer > single layer), the more significant the performance improvement.

\subsubsection{Normalized Attention Function}

The new attention function (NAF) proposed by KernelWarehouse optimizes the dynamic learning process of large-scale kernel unit mixing through three mechanisms. The core innovation of this function is the use of a linear normalization method instead of the traditional Softmax, dividing the original attention score $z_{ij}$ by the sum of the absolute values of all candidate kernel unit scores. This breaks the traditional attention's non-negative weight constraint, allowing negative values to participate in the mixing calculation to enhance the adversarial interaction between different kernel units. To alleviate the optimization difficulties of large-scale parameter mixing in the early stages of training, a temperature annealing mechanism is introduced: within the first 20 training epochs, the temperature coefficient $\tau$ linearly decays from 1 to 0, enabling the model to gradually transition from deterministic initialization to dynamic attention adjustment. During the initialization phase, a strong association between kernel units and mixing positions is established through predefined binary masks $\beta_{ij}$. When the parameter budget b$\geq$1, each mixing unit is forced to bind at least one exclusive kernel unit, while when b<1, each kernel unit is limited to serving at most one mixing position, ensuring the interpretability and training stability of the mixing system.

\begin{figure*}[ht!]  
  \centering
  \includegraphics[width=1.0\textwidth]{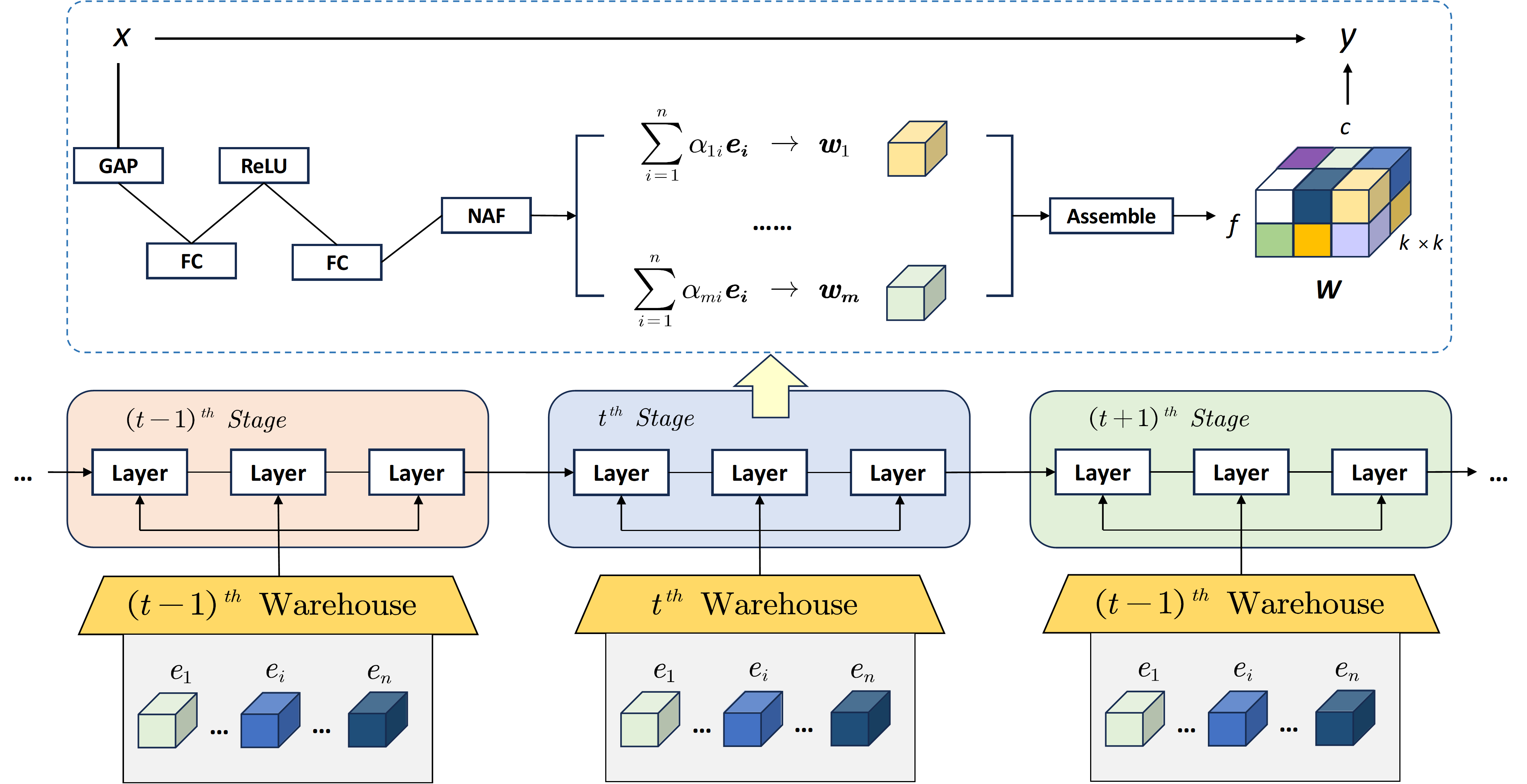}  
  \caption{Schematic illustration of KernelWarehouse.\label{fig3}}
\end{figure*} 

\subsection{Triple Attention Mechanism}

\begin{figure*}[ht!]  
  \centering
  \includegraphics[width=1.0\textwidth]{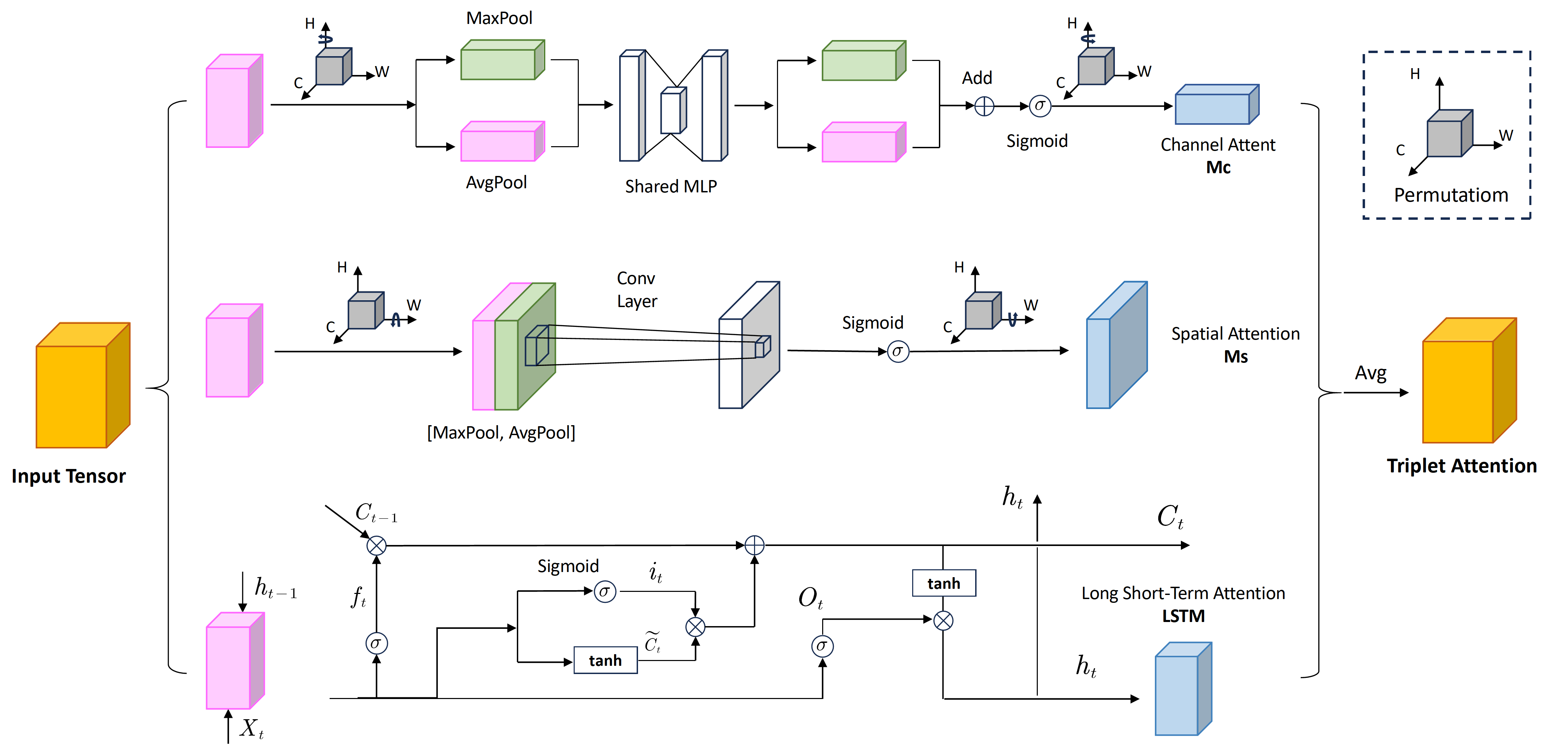}  
  \caption{Schematic illustration of Triple Attention Mechanism.\label{fig4}}
\end{figure*}

To address the diversity and morphological specificity of concrete crack detection targets, the network requires adaptive feature attention allocation that dynamically adjusts focus regions based on spatial distribution and geometric characteristics. This prevents information conflicts while maintaining robust feature learning across heterogeneous targets. Our refined Triple Attention (TA)\cite{35} mechanism, as depicted in Figure \ref{fig4}, specifically addresses scattered and multi-scale crack patterns through three parallel branches:

\begin{enumerate}
\item	\textbf{Spatial Attention Branch}: Prioritizes crack regions via 2D positional encoding;
\item	\textbf{Channel Attention Branch}: Amplifies crack-relevant feature channels using squeeze-excitation;
\item	\textbf{Cross-Dimensional Fusion Branch}: Harmonizes multi-scale features through learnable gating.
\end{enumerate}

Each branch independently captures inter-dimensional interactions while preserving parameter efficiency through shared kernel warehouses.

In the proposed network architecture, the first two branches focus on modeling interactions between channel (C) and spatial (H/W) dimensions, enabling precise adjustment of feature weights to prioritize critical crack characteristics. For dispersed targets such as low-density crack distributions, this channel-spatial attention mechanism enhances discriminative features while enriching contextual information from peripheral regions, effectively addressing dispersion challenges. The third branch integrates an LSTM-inspired design to establish long-range spatial attention, capturing contextual dependencies across extended spatial ranges (up to 100 pixels) for dynamic weight adjustment in distant regions\cite{36}.

Conventional triple attention mechanisms often exhibit localized bias, overemphasizing specific high-contrast regions while neglecting sparse or morphologically diverse cracks\cite{37}. The mechanism effectively enhances the detection of complex targets by jointly optimizing feature details and global information.

\subsubsection{First Branch: Channel Attention}

First, global max pooling and global average pooling are operated on the input feature map of size C×H×W to compress the spatial dimensions, resulting in two feature maps of size 1×1×C. Then, these two feature maps are input into a shared multilayer perceptron (MLP) to obtain two new feature maps of size 1×1×C. Finally, the outputs of the MLP are added together and processed through a Sigmoid activation function to generate the final channel attention weight matrix $M_c$. The formula for this is:

\begin{equation}
M_c\in R^{C\times 1\times 1}
\end{equation}

where the first layer of the MLP has C/r neurons with a ReLU activation function, and the second layer has C neurons. To reduce the computational parameters, the MLP introduces a dimensionality reduction factor r, which is represented by the formula:

\begin{equation}
M_c\in R^{C/r\times 1\times 1}
\end{equation}

Combining the above steps, the formula for channel attention can be summarized as:

\begin{equation}
M_c(F) = \sigma \Bigl( \text{MLP}\bigl( \text{AvgPool}(F) \bigr) + \text{MLP}\bigl( \text{MaxPool}(F) \bigr) \Bigr)\label{eq3}
\end{equation}

Using $F_{avg}^{c}$ and $F_{\max}^{c}$ to represent the global average pooling feature and the global max pooling feature respectively, the formula can be simplified as:

\begin{equation}  
M_c(F) = \sigma \Bigl( \text{MLP}\bigl(F_{\mathrm{avg}}^{c}\bigr) + \text{MLP}\bigl(F_{\max}^{c}\bigr) \Bigr)\label{eq4}
\end{equation}

The first branch decouples the processing of spatial and channel features while establishing cross-dimensional interactions. Spatial features, such as crack morphology and positional patterns, undergo coordinate-sensitive recalibration through deformable convolutions. Concurrently, channel features including texture gradients and chromatic characteristics are adaptively weighted via squeeze-excitation mechanisms. A cross-attention module synthesizes these decoupled representations by aligning spatial discontinuities with anomalous channel responses, prioritizing fracture-critical regions where material properties and geometric anomalies exhibit strong correlations. This joint optimization enables precise localization of crack boundaries while suppressing background interference from similar textures.

\subsubsection{Second Branch: Spatial Attention}

During the channel compression stage, the input feature tensor (with dimensions C×H×W) is first subjected to global maximum pooling and global average pooling. This reduces the dimensionality along the channel axis to generate two two-dimensional feature maps (each with size H×W×1). These two channel-compressed results are then concatenated along the channel axis to form a composite feature tensor containing dual spatial information (with dimensions H×W×2). Finally, after being processed by the Sigmoid function for nonlinear normalization, the spatial attention weight matrix $M_s$ is derived. Its mathematical representation is as follows:

\begin{equation}
M_s\left( F \right) \in R^{H,W}
\end{equation}

Similarly, the pooling methods in the spatial dimension generate two-dimensional feature maps, with the formulas for average pooling and max pooling being:

\begin{equation}
F_{\mathrm{avg}}^{s}\in R^{1\times H\times W}
\end{equation}
\begin{equation}
F_{\max}^{s}\in R^{1\times H\times W}
\end{equation}

Finally, based on the formulas for average pooling and max pooling, the formula for spatial attention can be derived as:

\begin{equation}
M_s\left( F \right) =\sigma \left( f^{7\times 7}\left[ F_{\mathrm{avg}}^{s};F_{\max}^{s} \right] \right)\label{eq8}
\end{equation}

The second branch enhances the first branch through differentiated weighting strategies for spatial and channel features, optimizing multi-scale feature representation. In scenarios with dispersed targets, spatial recalibration prioritizes target distribution patterns, while channel modulation adaptively emphasizes discriminative attributes. The synergistic interplay between these dimensions generates more discriminative feature embeddings by resolving spatial-channel misalignments in low-contrast regions.

\subsubsection{Third Branch: Temporal Attention}

The tertiary pathway incorporates an LSTM-inspired architecture comprising three operational phases. The forget gate mechanism selectively filters obsolete data elements from the memory state. Subsequently, the input gate governs the integration of novel information into the updated cellular state. Concurrently, the output gate modulates information propagation through sigmoidal gating operations, ensuring contextually relevant feature transmission.

\textbf{Forget Gate}: The forget gate takes the previous layer's output $h_{t-1}$ and the current layer's input sequence data $x_t$ as inputs. It passes these inputs through a sigmoid activation function to produce an output $f_t$, which controls the extent to which the previous layer's cell state $C_{t-1}$ is forgotten. The calculation formula is:

\begin{equation}
f_t=\sigma \left( W_f\cdot \left[ h_{t-1},x_t \right] +b_f \right)\label{eq9}
\end{equation}

\textbf{Input Gate}: 
The input gate incorporates dual computational stages: the initial stage employs a sigmoidal gating mechanism to regulate information flow, generating the input modulation factor $i_t$, while the subsequent stage utilizes hyperbolic tangent normalization to process state updates, yielding the candidate state vector $C_{t-1}$. These complementary operations are mathematically formalized as:
\begin{equation}
i_t=\sigma \left( W_i\cdot \left[ h_{t-1},x_t \right] +b_i \right)
\end{equation}
\begin{equation}
C_t=\tanh \!\:\left( W_c\cdot \left[ h_{t-1},x_t \right] +b_c \right)
\end{equation}

The product of $i_t\ast C_t$ represents the amount of new information to be retained. Together, these two gates determine how the new information updates the cell state $C_t$ of this layer, with the expression formula being:

\begin{equation}
C_t=f_t\ast C_{t-1}+i_t\ast C_t
\end{equation}

\textbf{Output Gate}: The output gate controls the amount of information to be filtered from the cell state of this layer. First, a value $o_t$ within the range [0,1] is obtained using the sigmoid activation function, with the expression formula being:

\begin{equation}
o_t=\sigma \left( W_o\left[ h_{t-1},x_t \right] +b_o \right)
\end{equation}

Finally, the cell state $C_t$ is processed through the tanh activation function and then multiplied with $o_t$ to obtain the output $h_t$ of this layer, with the expression formula being:

\begin{equation}
h_t=o_t\ast \tanh \!\:\left( C_t \right)\label{eq14}
\end{equation}

The third branch specializes in long-range spatial context modeling, distinct from the local feature weighting strategies of the first two branches. The LSTM-inspired spatial attention mechanism dynamically recalibrates spatial attention weights across extended receptive fields, enabling joint optimization of local discriminative features and global contextual cues. This proves particularly effective when targets appear distant from focal regions or blend with complex backgrounds, enhancing detection robustness in occluded scenarios.

The enhanced TA mechanism integrates LSTM-based sequential attention propagation with cross-dimensional interaction refinement, significantly improving detection performance for spatially dispersed and morphologically heterogeneous targets. Experimental validation demonstrates superior performance compared to conventional TA designs in high-clutter environments, particularly excelling in cases involving partial occlusion or low-contrast crack patterns.

\subsection{Loss Function}

Our proposed framework employs a composite loss architecture with dual optimization objectives: image segmentation and object detection. The segmentation component employs a modified cross-entropy criterion addressing class imbalance. Specifically, as crack pixels constitute minimal proportions in typical infrastructure imagery (often <5\% pixel coverage), conventional loss calculations become dominated by background regions. To address this inherent class imbalance, we implement a class-aware weighting mechanism within the pixel-wise cross-entropy framework\cite{25}. The enhanced weighted cross-entropy loss $L_{wce}$ is formulated as:

\begin{equation}
L_{wce}=-\frac{1}{N}\times \sum_{n=1}^N{\sum_{c=1}^2{w_c\times \log \frac{\exp \left( y_{n,c} \right)}{\exp \left( y_{n,i} \right)}}}\times y\prime_{n,c}
\end{equation}

where $y$ represents the input, $y\prime$ represents the target, $w$ represents the weight, and N represents the number of samples across the mini-batch dimension, which is set to 8 in this study.

The proposed FP-IoU loss tackles crack detection challenges in concrete imagery—edge blurring and surface complexity—by synergistically integrating Focaler IoU’s sample imbalance mitigation\cite{39} and PIoUv2’s anchor dilation correction\cite{40}. Replacing static penalties\cite{38} with adaptive interval mapping, it suppresses regression-induced box expansion while applying difficulty-aware weighting to low-contrast cracks. This dual mechanism accelerates convergence and enhances precision through defect-aligned gradient updates.

The Focaler IoU function is shown in Equation \ref{eq16}, which constructs IoU in the form of a piecewise function to improve the regression effect of bounding boxes.

\begin{equation}
L_{\mathrm{IoU focaler}}=\left\{ \begin{array}{c}
	0,\mathrm{}L_{\mathrm{IoU}}<d\\
	\frac{L_{\mathrm{IoU}}-d}{u-d},\mathrm{}d\leqslant L_{\mathrm{IoU}}\leqslant u\\
	1,\mathrm{}L_{\mathrm{IoU}}>u\\
\end{array} \right.\label{eq16}
\end{equation}

In the equation, $L_{\mathrm{IoU}}$ refers to the original value, and d and u are both within the interval of 0 to 1. Different regression samples correspond to different d and u. The loss is defined as shown in Equation \ref{eq17}:

\begin{equation}
X_{\mathrm{IoUfocaler}}=1-L_{\mathrm{IoUfocaler}}\label{eq17}
\end{equation}

By examining the Focaler IoU formula, it can be seen that when $L_{\mathrm{IoU}}$ is less than a specific value, $L_{\mathrm{IoUfocaler}}$ is zero. However, the research object of this paper is concrete cracks, which are mostly low-quality target defects and do not match the original loss function. The improved loss function adopts a piecewise function form, eliminating the original parameters, which allows for more efficient use of sample data, better handling of low-quality defects, comprehensive analysis of sample positions, and promotes the stable convergence of the model. This enhances the performance on the dataset of this paper. The loss definitions are given in Equations \ref{eq18}-\ref{eq19}:

\begin{equation}
L_{\mathrm{IoU}}=\left\{ \begin{array}{c}
	\frac{L_{\mathrm{IoU}}}{u},\mathrm{}L_{\mathrm{IoU}}\leqslant u\\
	1,\mathrm{}L_{\mathrm{IoU}}>u\\
\end{array} \right.\label{eq18}
\end{equation}
\begin{equation}
X_{\mathrm{IoU}^{\mathrm{f}}}=1-L_{\mathrm{IoU}^{\mathrm{f}}}\label{eq19}
\end{equation}

\begin{figure}[t!]
  \centering
\includegraphics[width=7cm]{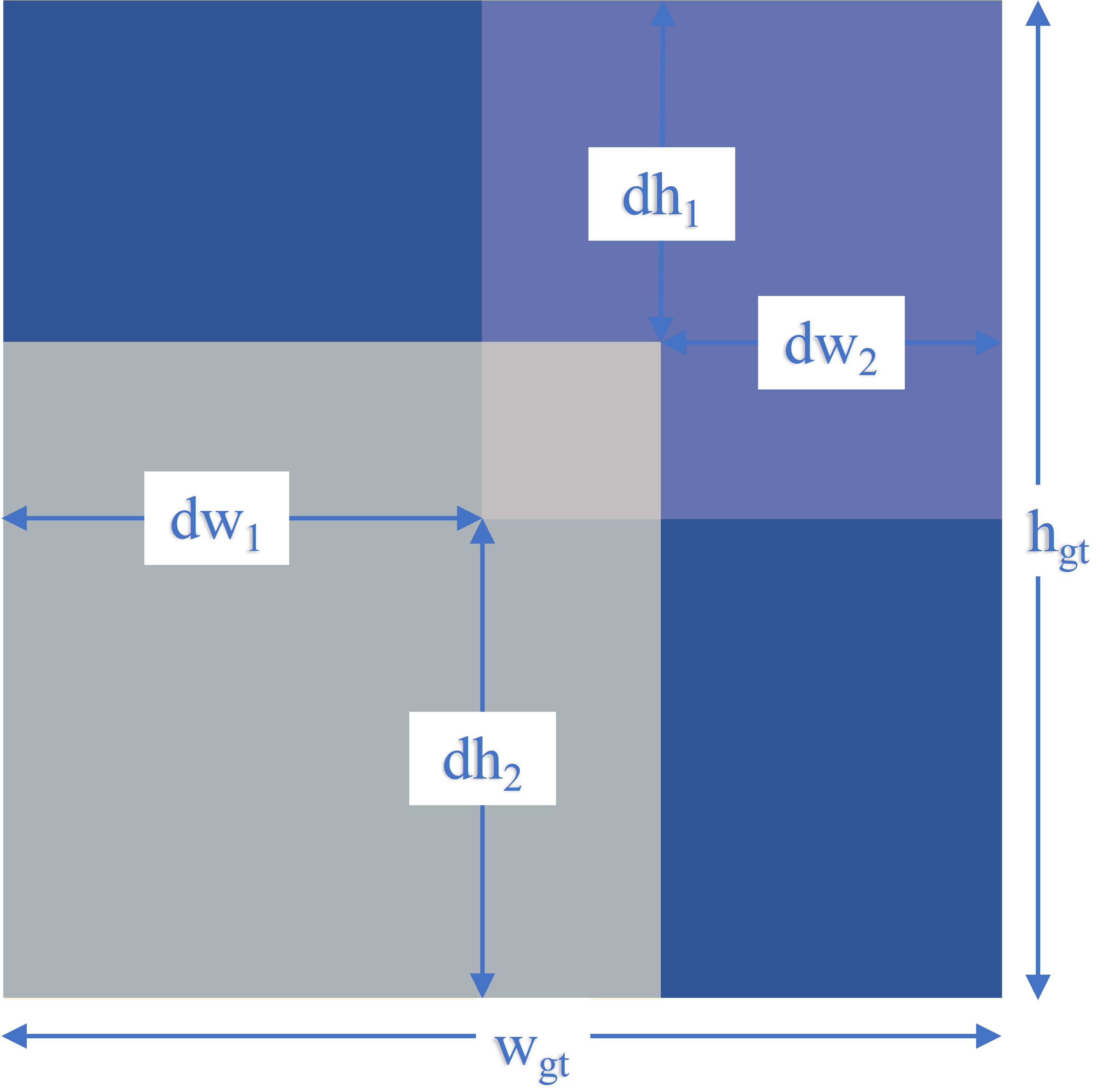} 
\caption{Variables defined in PIoU.\label{fig5}}
\end{figure}

PIoU loss addresses inefficiencies in conventional anchor box regression by introducing adaptive penalty factors and gradient reshaping to optimize update trajectories. Instead of inducing sequential expansion-shrink cycles, it directly minimizes inter-edge distances between anchors and ground-truth boxes (Equations \ref{eq20}-\ref{eq21}). This dual mechanism enables near-linear regression paths through quality-aware gradient modulation, accelerating convergence while maintaining aspect ratio consistency—particularly crucial for slender targets like cracks. By dynamically adjusting penalties based on target dimensions and spatial context, PIoU bypasses suboptimal intermediate states, achieving precise localization through geometrically coherent updates.

\begin{equation}
p = \text{\small $\frac{1}{4}$} \left( \frac{dw_1}{w_{\mathrm{gt}}} + \frac{dw_2}{w_{\mathrm{gt}}} + \frac{dh_1}{h_{\mathrm{gt}}} + \frac{dh_2}{h_{\mathrm{gt}}} \right) \label{eq20}
\end{equation}
\begin{equation}
X_{\mathrm{PIoU}}=L_{\mathrm{IoU}}+1-\mathrm{e}^{-p^2}\label{eq21}
\end{equation}

where P is the penalty factor, and the defined variables are shown in Figure \ref{fig5}.

Building upon the PIoU framework, a non-monotonic attention layer m(x) is introduced to refine the focus mechanism through dynamic optimization weight adjustment. By integrating this attention layer with PIoU, the PIoUv2 loss function is developed to prioritize medium-quality anchor boxes that exhibit ambiguous regression characteristics. Unlike its predecessor, PIoUv2 employs quality-aware gradient modulation to balance optimization efforts between high-quality anchors (precise localization) and low-quality outliers (robust error suppression), effectively addressing the oversight of transitional-state samples in standard implementations. This dual-balancing mechanism, mathematically formalized in Equations \ref{eq22}-\ref{eq24}, enhances detection stability for edge-blurred cracks by adaptively recalibrating penalty intensities based on anchor box positioning confidence and spatial coherence with ground-truth boundaries.

\begin{equation}
q=\mathrm{e}^{-p},q\in (0,1]\label{eq22}
\end{equation}
\begin{equation}
m\left( x \right) =3x\cdot \mathrm{e}^{-x^2}\label{eq23}
\end{equation}
\begin{equation}
X_{\mathrm{PIoUv}2\mathrm{}}=3m(\lambda q)\cdot X_{\mathrm{PIoU}}\label{eq24}
\end{equation}

From Equation \ref{eq22}, an increase in p corresponds to a decrease in q. q signifies the quality of an anchor box. When p is 0, q equals 1, indicating a perfect overlap between the ground-truth box and the predicted box at that moment. $\lambda$ is a hyperparameter that regulates the attention.

This paper proposes the FP-IoU loss function by integrating the Focaler IoU and PIoUv2 loss functions. The definition of the loss is as follows:

\begin{equation}
X_{\mathrm{FP}-\mathrm{IoU}}=3m(\lambda q)\left( X_{\mathrm{IoU}^{\mathrm{f}}}+1-\mathrm{e}^{-p^2} \right)
\end{equation}

The FP-IoU loss function proposed in this study synergistically integrates design principles from two distinct loss function paradigms. By dynamically modulating penalty factors in proportion to target dimensions\cite{38} and implementing a holistic positional encoding strategy that encapsulates both edge-level geometric relationships and region-wise spatial coherence\cite{39}, FP-IoU establishes a quality-aware gradient redistribution mechanism. This integrated approach enables simultaneous optimization of spatial-spectral feature disentanglement for high-quality anchors while suppressing error propagation from low-confidence proposals. Compared to CIoU variants, FP-IoU introduces three key enhancements: adaptive curvature adjustment based on crack width/length ratios, multi-scale penalty factor fusion through attention-weighted feature pyramids, and dynamic gradient clipping thresholds to prevent over-penalization of partially occluded targets. Tests show our dual-path design achieves higher accuracy than standard CIoU loss, with stable performance on both dense and sparse cracks.

\section{Experiment and Results}
\subsection{Datasets}
\begin{figure*}[t!]  
  \centering
  \includegraphics[width=1.0\textwidth]{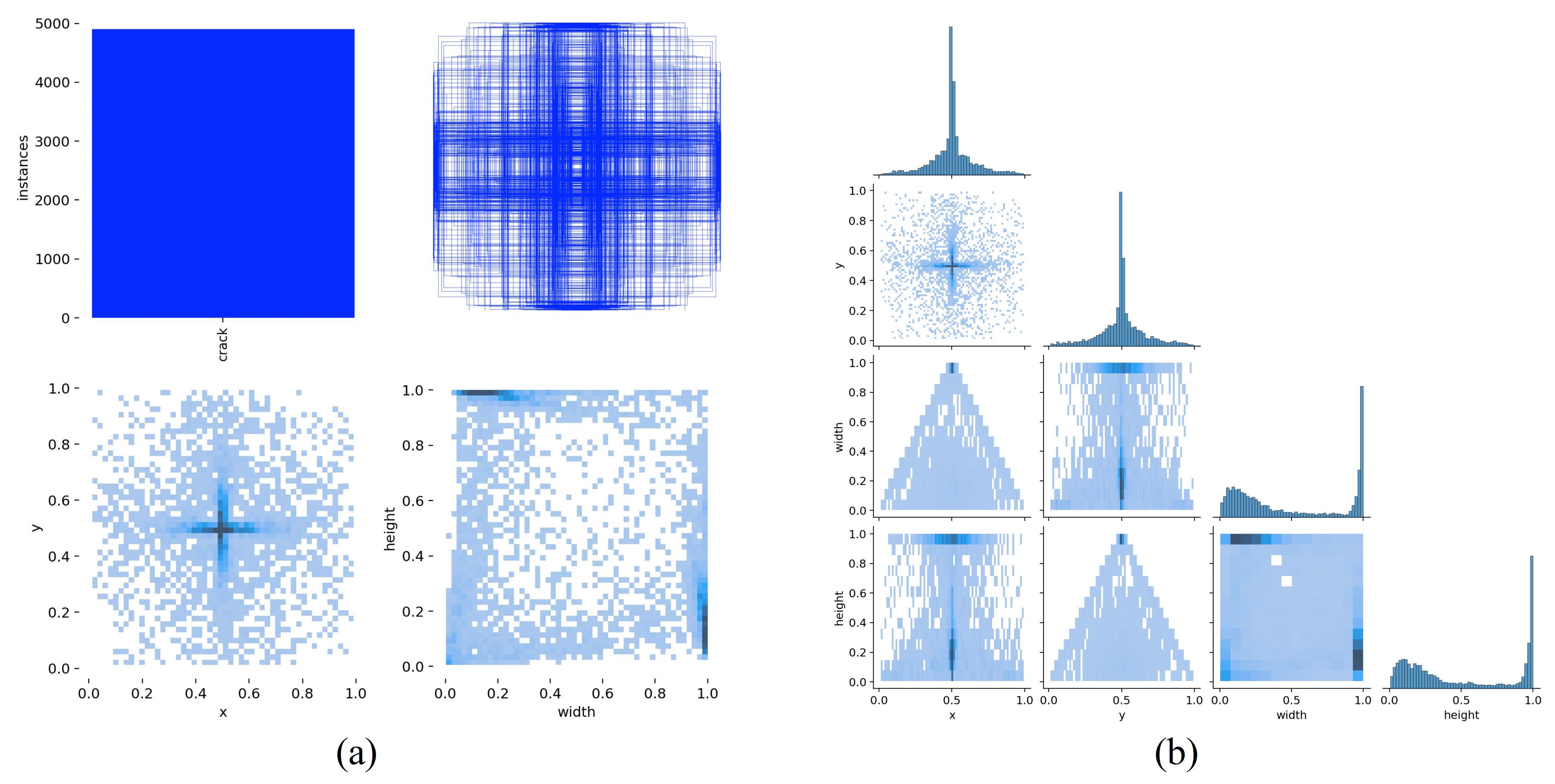}  
  \caption{Visualization of the detailed analysis of the Crack-Seg dataset; (a)labels; (b)labels-correlogram.\label{fig6}}
\end{figure*}

To ensure that the YOLOv11-KW-TA-FP model has high generalization ability and performance, this paper selects three public datasets for experiments, namely the Crack-Seg dataset, the Surface Crack Detection dataset, and the Crack Segmentation dataset. Among them, the Crack-Seg dataset is mainly used in the model training stage. The Surface Crack Detection dataset and the Crack Segmentation dataset are mainly used in the generalization experiment part. To ensure the reliability of the model training process and the generalization ability of the model, the above three datasets have been preprocessed. First, we use the perceptual hash algorithm to calculate the 64-bit hash fingerprint of each image and set the Hamming distance threshold to 5 for similarity detection. For the detected duplicate images, only the earliest sample is retained, and the rest are removed from the dataset. Then, according to the standard data allocation protocol, the Crack-Seg dataset is divided into training, validation, and test sets in the ratio of 7:2:1. The detailed analysis of the Crack-Seg dataset is shown in Figure \ref{fig6}. In (a), from left to right, grid 1 shows the data volume of the training set, displaying the number of samples contained in each category. Grid 2 shows the size and number of boxes, displaying the size distribution and corresponding number of bounding boxes in the training set. Grid 3 shows the position of the center point relative to the entire image, describing the distribution of the center points of the bounding boxes in the image. Grid 4 shows the height-to-width ratio of the target in the image relative to the entire image, reflecting the distribution of the height-to-width ratio of the targets in the training set. (b) shows the modeling of the correlation between labels by the object detection algorithm during the training process. Each matrix cell represents the label used in model training, and the color depth of the cell reflects the correlation between the corresponding labels.

\subsection{Experimental Settings}
\unskip
\subsubsection{Experimental Environment and Hyperparameter Settings}

The experiment is based on the Windows 10 22H2 operating system, using a 12th Gen Intel(R) Core(TM) i5-12600KF CPU, an Nvidia RTX 4070Ti Super 16G GPU, the deep learning framework PyTorch version 1.13.1, a Python environment of 3.12.2, and CUDA version 12.6.The experimental hyperparameter settings are shown in table \ref{tab1}.

\begin{table}[!ht]
\caption{Hyperparameter Settings.\label{tab1}}
\centering
\renewcommand{\arraystretch}{1} 
\begin{tabularx}{\columnwidth}{*{6}{>{\centering\arraybackslash}X}}
\toprule
\textbf{Epoch} & \textbf{Batch Size} & \textbf{Optimizer} & \textbf{Momentum} & \textbf{Ir0} & \textbf{Image Size} \\ 
\midrule
200 & 16 & SGD & 0.937 & 0.01 & 640×640 \\ 
\bottomrule
\end{tabularx}
\end{table}

\subsubsection{Model Evaluation Metrics}

To systematically validate the model's comprehensive performance, this study employs an integrated evaluation framework combining qualitative visual analysis and quantitative metric benchmarking. The qualitative assessment involves visual comparisons between detection results from the proposed YOLOv11-KW-TA-FP model and state-of-the-art baselines, with focused analysis on three critical aspects: localization precision of crack boundaries, consistency in detecting multi-scale cracks, and reduction rates of false positives (e.g., misclassified concrete textures) and false negatives (e.g., overlooked hairline fractures). This visualization-driven approach provides intuitive insights into the model's capability to handle edge-blurred cracks under complex background interference.

Quantitative evaluation adopts standard object detection metrics including precision (P), recall (R), and mean average precision (mAP). The mAP metric is further decomposed into mAP 50(calculated at an FP-IoU threshold of 0.5) and mAP 50:95(averaged across FP-IoU thresholds from 0.5 to 0.95 with 0.05 increments), as mathematically formalized in Equations \ref{eq26}-\ref{eq29}. These metrics systematically quantify the model's ability to balance detection sensitivity (minimizing missed cracks) and specificity (reducing false alarms), particularly critical for infrastructure inspection scenarios where over-detection could lead to unnecessary maintenance costs. By integrating visual interpretability with statistical rigor, this dual evaluation paradigm ensures reliable performance validation for field-deployable crack detection systems.

\begin{equation}
P=\frac{T_P}{T_P+F_P}\label{eq26}
\end{equation}
\begin{equation}
R=\frac{T_P}{T_P+F_N}\label{eq27}
\end{equation}

In Equation \ref{eq26}, TP denotes the number of true positive samples detected, and FP denotes the number of false positive samples predicted. In Equation \ref{eq27}, FN represents the number of positive samples that were not detected.

\begin{equation}
AP=\int_0^1{P\cdot RdR}\label{eq28}
\end{equation}
\begin{equation}
mAP=\frac{\sum_{i=0}^n{AP_i}}{n}\label{eq29}
\end{equation}

In Equation \ref{eq29}, n denotes the number of detected object categories.

\subsection{Results and Disscussions}
\unskip
\subsubsection{Experimental Results of YOLOv11-KW-TA-FP model}

\begin{figure}[t!]
\centering
\includegraphics[width=0.8\linewidth]{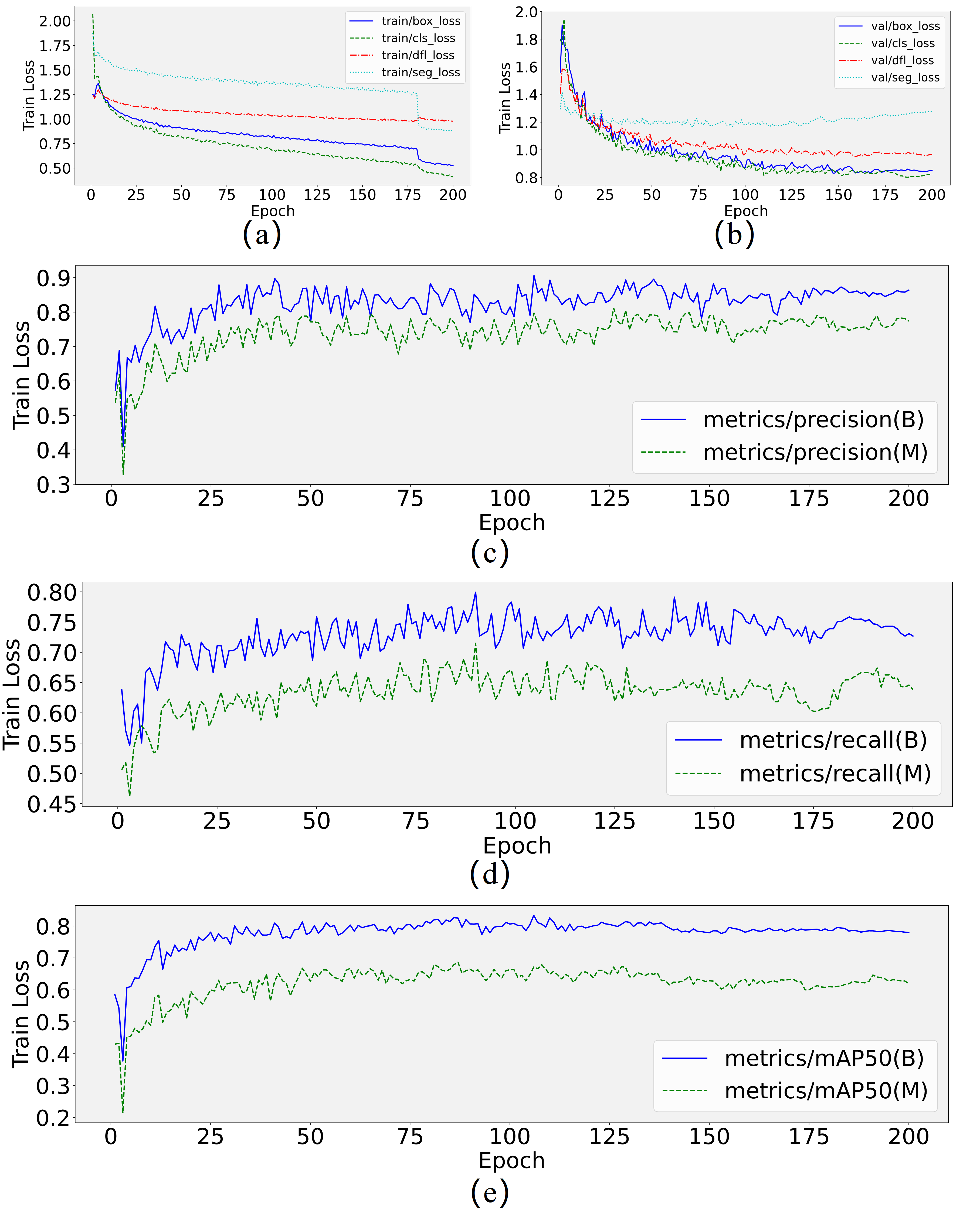} 
\caption{Model training results (B and M represent Box and Mask respectively).
(a) Box loss, classification loss, DFL loss, and segmentation loss curves on the training dataset;
(b) Box loss, classification loss, DFL loss, and segmentation loss curves on the validation dataset;
(c) Precision curve during training;
(d) Recall curve during training;
(e) mAP@50 curve during training.\label{fig7}}
\end{figure}

The evaluation metrics of the developed YOLOv11-KW-TA-FP model all performed well. This section presents the experimental results of the YOLOv11-KW-TA-FP model after 200 training epochs, as shown in Figure \ref{fig7}. It displays the loss curves, precision curves, recall curves, and mAP50 curves of the training dataset and validation dataset. It can be seen that the model began to converge significantly after 25 training epochs. After 180 epochs with mosaic augmentation turned off, the curves remained stable, demonstrating good robustness.

\begin{figure}[t!]
\includegraphics[width=\columnwidth]{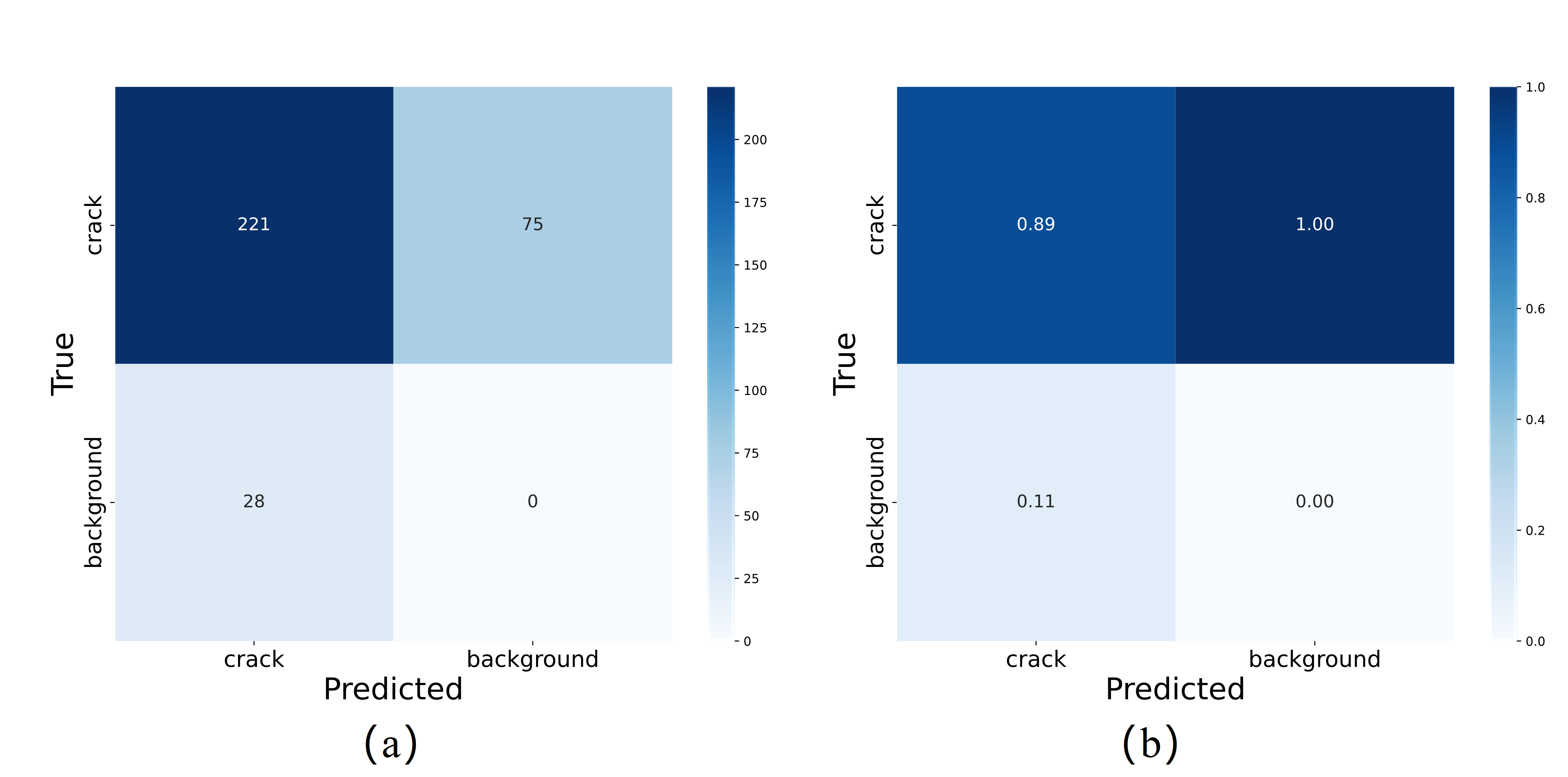} 
\caption{Confusion Matrix of the Developed YOLOv11-KW-TA-FP Model. (a) Confusion matrix; (b) Normalized confusion matrix.\label{fig8}}
\end{figure}

To visually demonstrate the model's classification performance, we visualized the confusion matrix of the YOLOv11-KW-TA-FP model's prediction results. As shown in Figure \ref{fig8}.The model has a high precision rate of 89\% in identifying crack and non-crack areas, indicating that it can correctly recognize most cracks and background areas. The relatively small number of background areas misidentified as cracks shows that the model performs well in terms of false positives and false negatives. The visualization of the confusion matrix further helps to understand the model's classification performance and provides references for optimization and improvement.The initial experimental analysis indicates that the model has a misclassification phenomenon for background-class samples. Specifically, the results shown in Figure 8 indicate that the model incorrectly classified 28 background images as crack images. After in-depth analysis, this issue is attributed to two key factors: (1) the insufficient diversity of background samples in the training dataset; (2) the loss function design lacks sufficient regularization constraints for background regions. To enhance the model's discrimination ability, future research plans to explore the introduction of a background penalty weight mechanism or the use of data augmentation techniques to enrich the diversity of training backgrounds.

\begin{figure*}[t!]  
  \centering
  \includegraphics[width=1.0\textwidth]{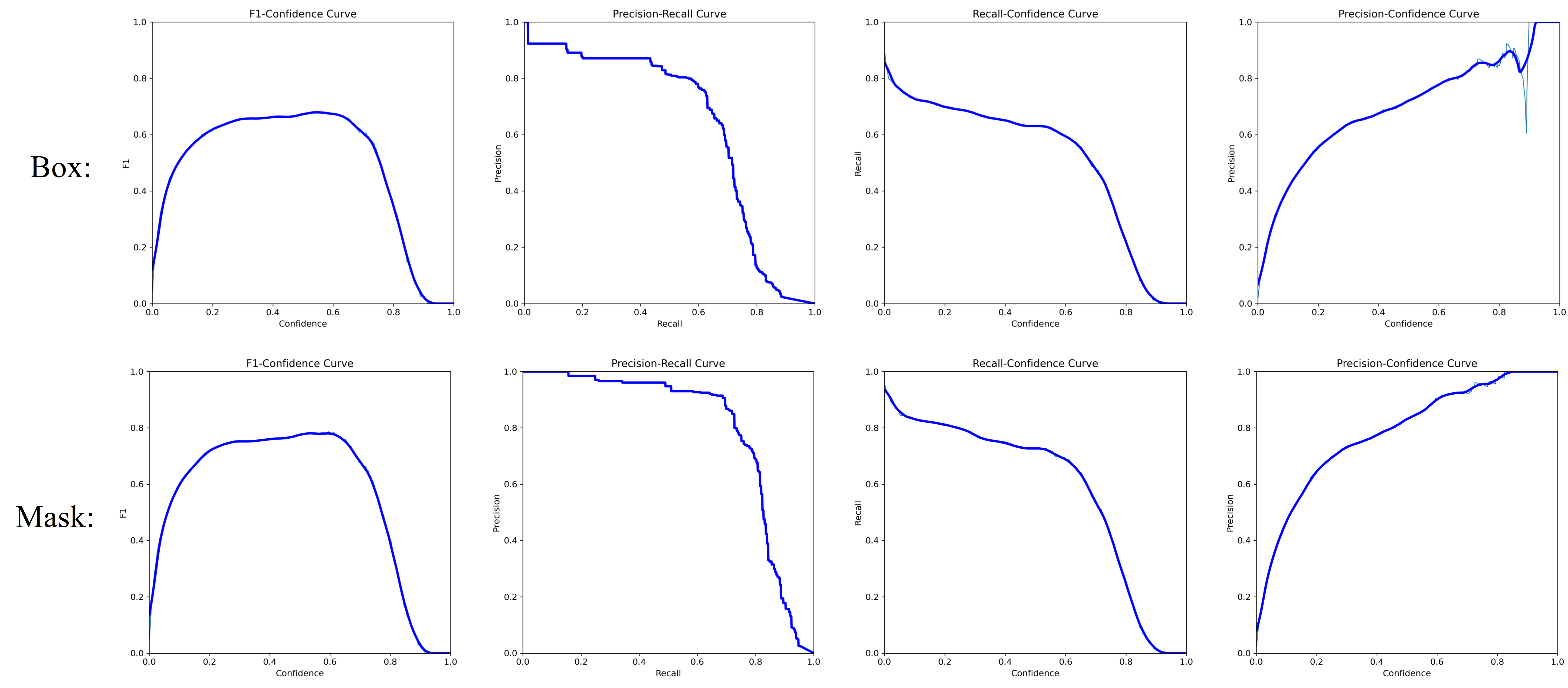}  
  \caption{The evaluation metric curves for detection and segmentation tasks.\label{fig9}}
\end{figure*}

Figure \ref{fig9} records the evaluation metric curves of the model in detection and segmentation tasks during training, including the F1 confidence curve, PR curve, recall confidence curve, and precision curve. The first row in Figure \ref{fig9} shows the evaluation metric curves for the object detection task, while the second row shows the evaluation metric curves for the segmentation task.

\subsubsection{Comparative Experiments of Different YOLOv11-KW-TA-FP Models}

\begin{figure}[t!]
\includegraphics[width=\columnwidth]{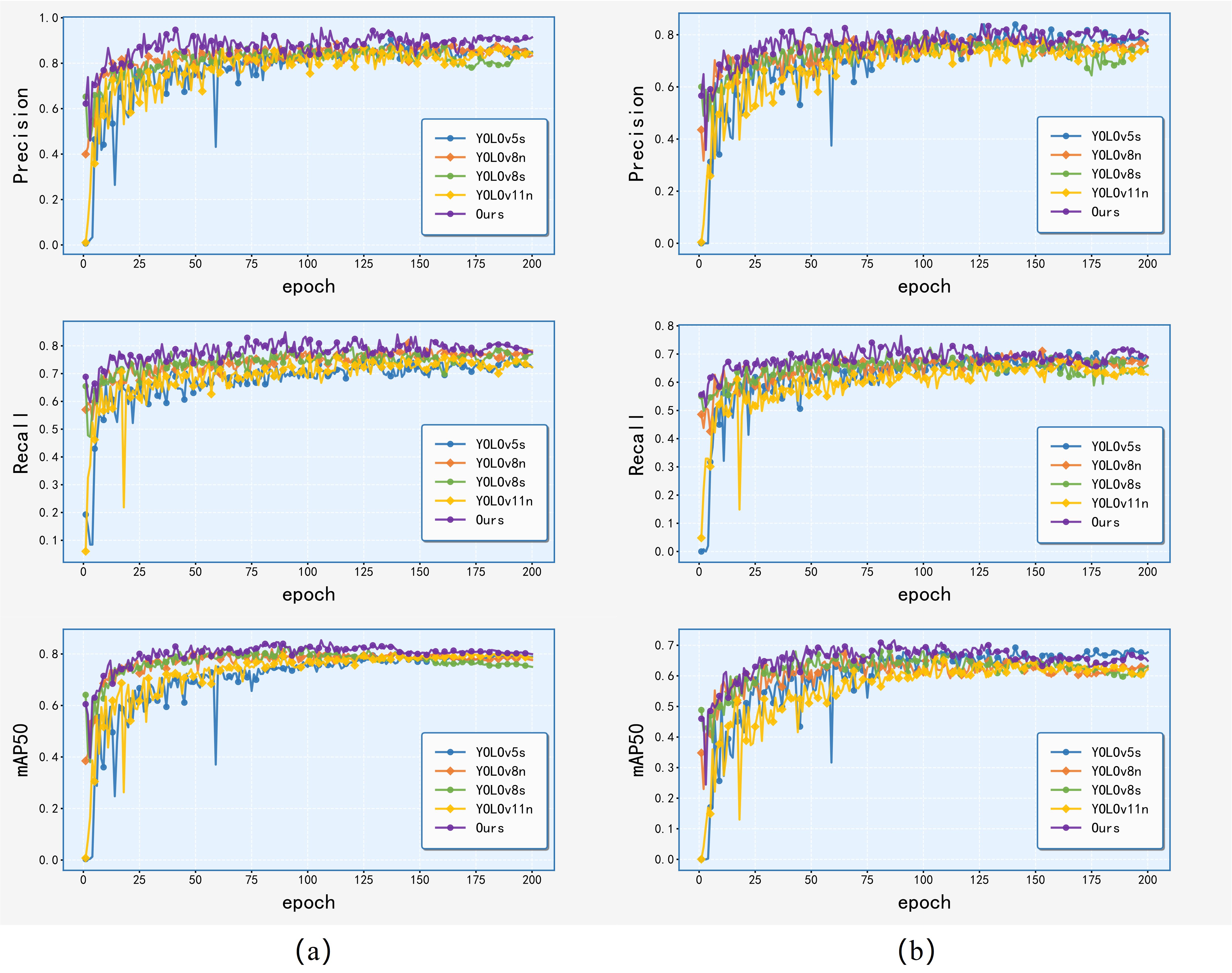} 
\caption{Evaluation metrics for the two types of tasks.(a) Evaluation metrics for object detection; (b) Evaluation metrics for segmentation tasks.\label{fig10}}
\end{figure}

The YOLOv11-KW-TA-FP model proposed in this paper demonstrates significant advantages in both concrete crack detection and segmentation tasks. As shown in Table 2, in the crack detection task, the model achieves a precision (P) of 91.3\% and an mAP50 of 86.4\%, surpassing all YOLO series models. Compared with DDBNet\cite{41}, CrackFormer\cite{42}, SSD\cite{43}, and Faster R-CNN\cite{44} models, the mAP50 is increased by 9.9\%, 14\%, 8.8\%, and 8.7\%, respectively. In the segmentation task, the model leads most models with a precision of 86.2\% and an mAP50 of 76.3\%, which is 9.2\% and 10.9\% higher than the baseline model YOLOv11n. Meanwhile, compared with FCN\cite{45} and Deeplabv3\cite{46} models that focus on segmentation tasks, the mAP50 is increased by 4\% and 2.5\%, respectively. The results indicate that the dynamic convolution (KWConv) optimizes the extraction of multi-scale crack features, the triple attention (TA) effectively suppresses complex background interference, and the FP-IoU loss function significantly improves the regression precision of low-quality samples. The combined effect of these three factors enables the model to achieve a balance between precision and generalization ability while maintaining a high recall rate. In addition, to demonstrate the advantages of the YOLOv11-KW-TA-FP model over other YOLO series models during training, this paper visualizes the evaluation metrics of precision, recall, and mAP50 for each training round of the YOLOv5s, YOLOv8n, YOLOv8s, YOLOv11n, and YOLOv11-KW-TA-FP models, as shown in Figure \ref{fig10}.

\begin{table*}[!ht]
\caption{Comparison of evaluation metrics across different models.\label{tab2}}
\centering
\renewcommand{\arraystretch}{1} 
\begin{tabularx}{\textwidth}{>{\centering\arraybackslash}X *{6}{>{\centering\arraybackslash}X}} 
\toprule
\multirow{2}{*}{\textbf{Network}} & \multicolumn{3}{c}{\textbf{Evaluation metrics of crack detection}} & \multicolumn{3}{c}{\textbf{Evaluation metrics of crack segmentation}} \\
\cmidrule(lr){2-4} \cmidrule(lr){5-7}
& P\%  & R\%  & mAP50\% & P\%  & R\%  & mAP50\% \\
\midrule
 YOLOv5n       & 85.3 & 72.2 & 78.2    & 74.3 & 68.7 & 66.4 \\
            YOLOv5s       & 86.6 & 70.0 & 79.7    & 75.0 & 62.7 & 63.3 \\
            YOLOv8n       & 85.8 & 72.9 & 78.7    & 81.0 & 67.1 & 71.2 \\
            YOLOv8s       & 85.8 & 73.5 & 79.6    & 83.5 & 68.3 & 71.5 \\
            YOLOv11n      & 87.1 & 73.3 & 79.2    & 77.0 & 69.9 & 65.4 \\
            SSD           & 86.5 & 72.3 & 77.6    & 73.2 & 63.8 & 64.3 \\
            Faster R-CNN  & 87.3 & 72.8 & 77.7    & 68.3 & 62.3 & 63.8 \\   
            FCN           & $-$  & $-$  & $-$     & 82.8 & 69.5 & 72.3 \\  
            Deeplabv3     & $-$  & $-$  & $-$     & 83.2 & 70.1 & 73.8 \\            
            DDBNet        & 68.7 & 63.9 & 76.5    & 82.1 & 70.2 & 73.1 \\
            CrackFormer   & 82.2 & 69.3 & 72.4    & 84.3 & 69.2 & 72.6 \\
            \textbf{Ours} & \textbf{91.3} & \textbf{76.6} & \textbf{86.4}    & \textbf{86.2} & \textbf{72.6} & \textbf{76.3} \\
\bottomrule
\end{tabularx}
\end{table*}

To comprehensively assess the overall performance of the improved model, Table \ref{taba} provides a comprehensive comparison of YOLOv11-KW-TA-FP with mainstream object detection/segmentation models in terms of crack detection accuracy, segmentation performance, and computational efficiency. The evaluation metrics for object detection tasks include FPS, Latency, and GFLOPs; the evaluation metric for segmentation tasks is Inference time. Finally, the number of parameters for all models is provided.

\begin{table*}[!ht]
\caption{Comprehensive comparison of performance across different Models.\label{taba}}
\centering
\renewcommand{\arraystretch}{1}
\begin{tabularx}{\textwidth}{
    >{\centering\arraybackslash}X
    >{\centering\arraybackslash}X
    >{\centering\arraybackslash}X
    >{\centering\arraybackslash}X
    >{\centering\arraybackslash}X
    >{\centering\arraybackslash}X
}
\toprule
\multirow{2}{*}{\textbf{Network}}
& \multicolumn{3}{c}{\textbf{Evaluation metrics of crack detection}}
& \textbf{Segmentation}
& \multirow{2}{*}{\textbf{Params(M)}} \\
\cmidrule(lr){2-4} \cmidrule(lr){5-5}
& \textbf{FPS (f/s)} & \textbf{Latency (ms)} & \textbf{GFLOPs}
& \textbf{Inference(ms)}
& \\
\midrule
YOLOv5n       & 81    & 1.62   & 4.2    & 98.3   & 2.18  \\
YOLOv5s       & 96    & 2.11   & 15.8   & 78.6   & 7.23  \\
YOLOv8n       & 85    & 1.35   & 8.1    & 74.8   & 3.01  \\
YOLOv8s       & 113   & 2.87   & 28.8   & 63.2   & 11.23 \\
YOLOv11n      & 88    & 1.36   & 6.7    & 45.3   & 2.66  \\
SSD           & 63    & 12.13  & 121.3  & 383.4  & 28.67 \\
Faster R-CNN  & 18    & 123.84 & 258    & 180.5  & 45.32 \\   
FCN           & $-$   & $-$    & $-$    & 5682.2 & 134.86 \\  
Deeplabv3     & $-$   & $-$    & $-$    & 3220.7 & 48.32 \\            
DDBNet        & 36    & 36.81  & 58.5   & 93.2   & 18.37 \\
CrackFormer   & 24    & 64.32  & 145.6  & 63.8   & 78.29 \\
\textbf{Ours} & \textbf{128} & \textbf{3.21} & \textbf{8.8} & \textbf{58.2} & \textbf{5.83} \\
\bottomrule
\end{tabularx}
\end{table*}

\begin{figure}[t!]
\includegraphics[width=\columnwidth]{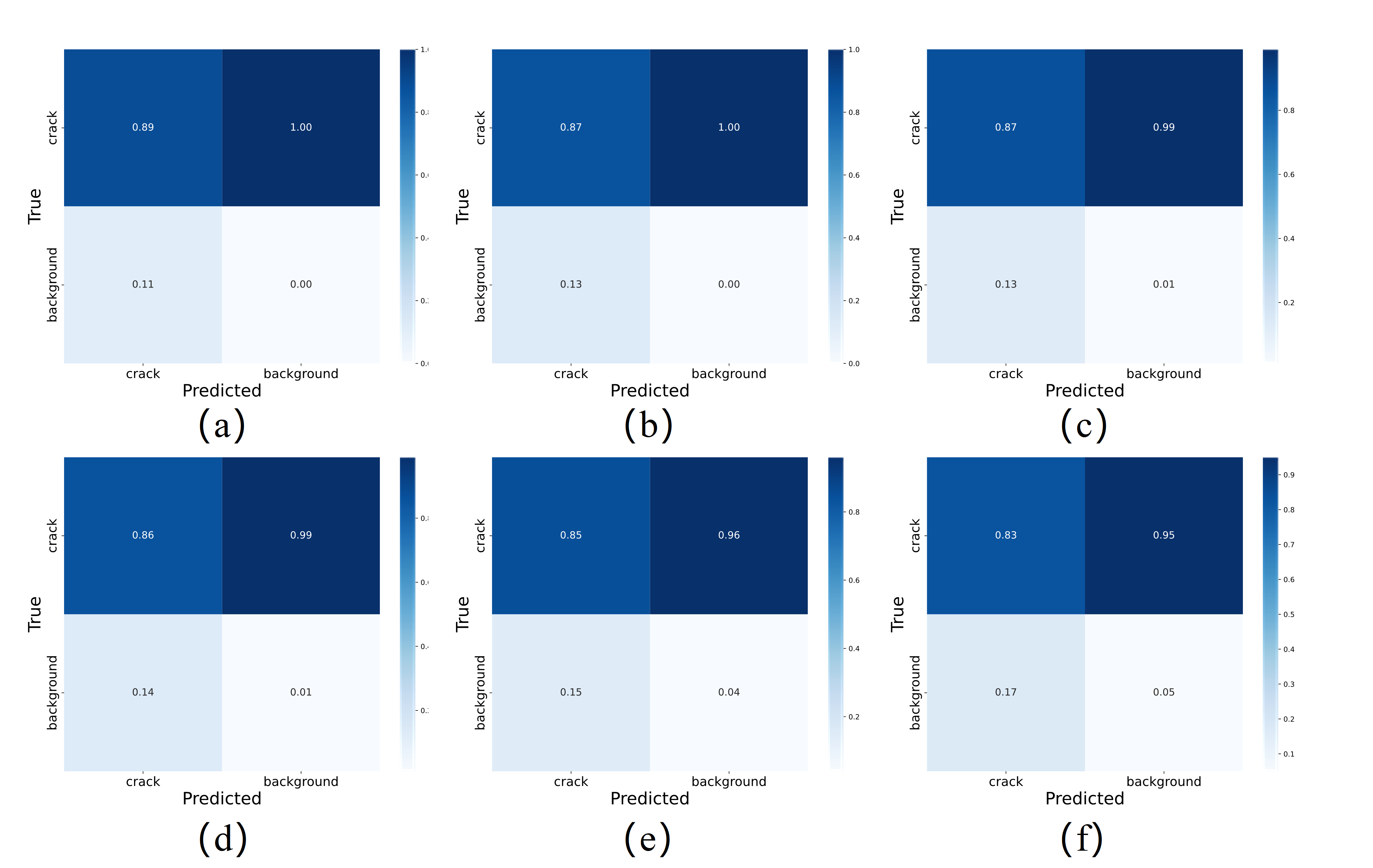} 
\caption{Comparison of confusion matrices. (a) YOLOv11-KW-TA-FP; (b) YOLOv11n; (c) YOLOv8s; (d) YOLOv8n; (e) YOLOv5s; (f) YOLOv5n.\label{fig11}}
\end{figure}

The localization performance of the YOLOv11-KW-TA-FP model is validated through confusion matrix analysis in Figure \ref{fig11}, which compares the validation set results of YOLOv5n, YOLOv5s, YOLOv8n, YOLOv8s, YOLOv11n, and the proposed model. The improved model exhibits higher values along the main diagonal and lower values on the secondary diagonal compared to other architectures, demonstrating enhanced capability in precisely localizing bridge cracks while minimizing misclassification errors. This pattern confirms the model's superior performance in crack localization tasks.

\begin{figure}[t!]
\includegraphics[width=\columnwidth]{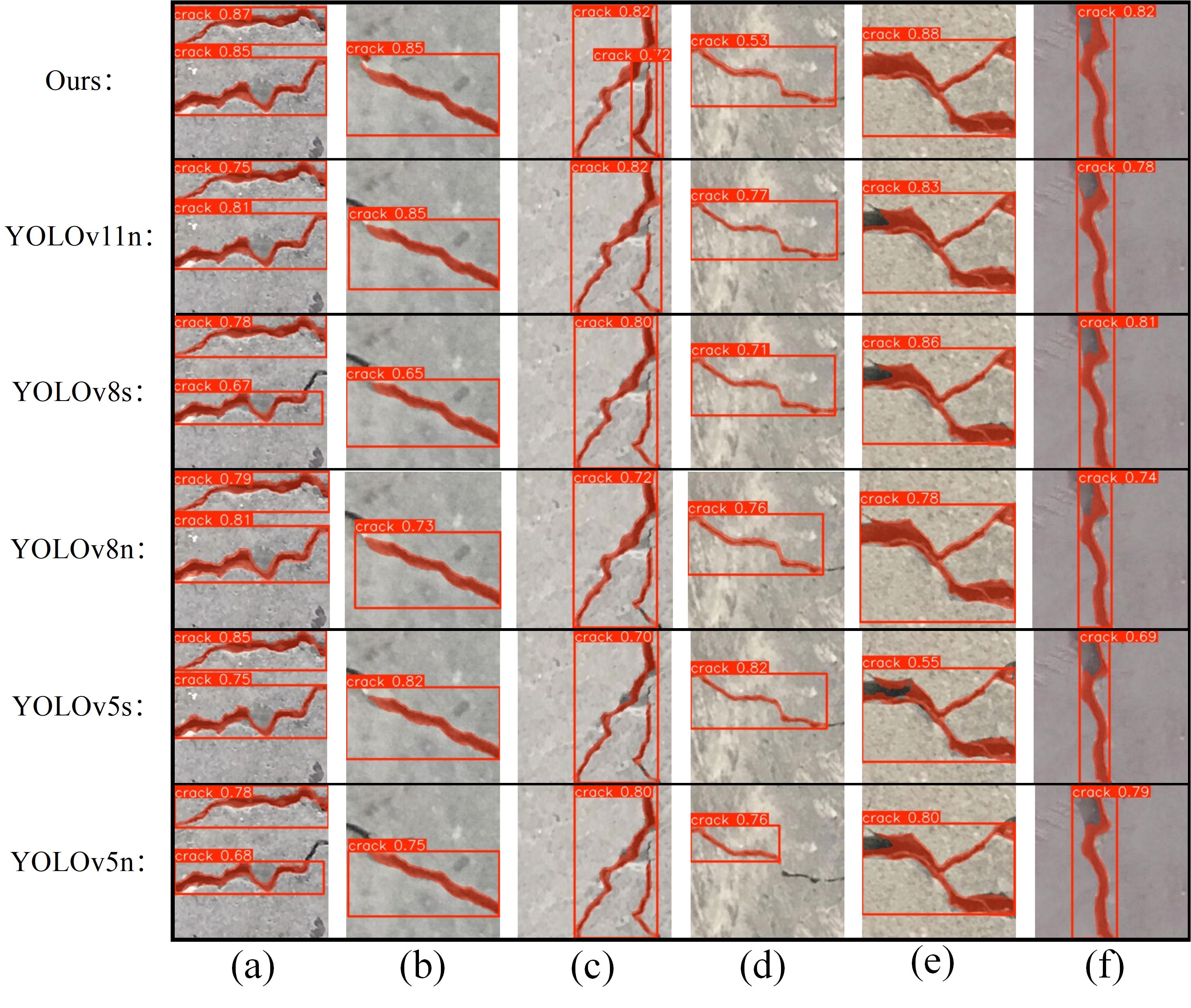} 
\caption{Model Result Comparison. (a)-(f) represent six different sets of images.\label{fig12}}
\end{figure}

To more intuitively demonstrate the performance of different models in the task of concrete crack detection and segmentation, we selected six representative images from the Surface Crack Detection dataset for testing and compared the detection results of each model. The results are shown in Figure \ref{fig12}.

In the concrete crack detection task, the improved YOLOv11-KW-TA-FP model achieves the highest confidence scores across image groups (a), (b), (c), (e), and (f) compared to other models. For image group (d), it demonstrates the most precise boundary delineation of crack regions. In segmentation tasks, image group (e) reveals that the model produces the finest segmentation details, accurately extracting complete crack structures from complex backgrounds.

\begin{figure}[t!]
\includegraphics[width=\columnwidth]{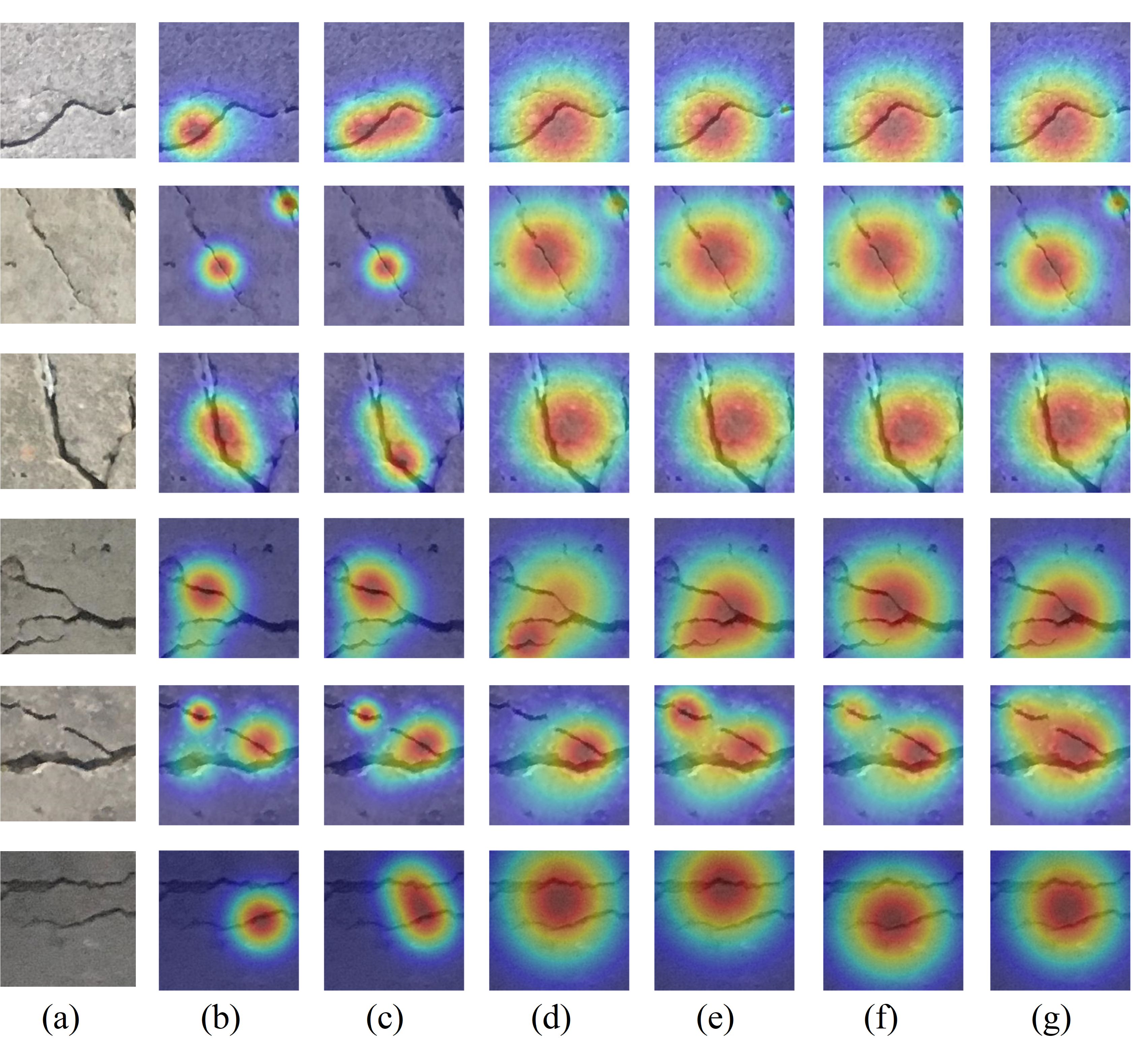} 
\caption{Comparison of Heatmaps for Different YOLO Model Series.(a) Original image, (b) YOLOv5n; (c) YOLOv5s; (d) YOLOv8n; (e) YOLOv8s; (f) YOLOv11; (g) YOLOv11-KW-TA-FP.\label{fig13}}
\end{figure}

To intuitively demonstrate the impact weight distribution of different rock feature regions on the results of recognition and segmentation tasks, this study employs the Grad-CAM feature visualization technique to generate heatmaps for six crack samples in the validation set, thereby comparing different series of YOLO models. This visualization technique can effectively highlight the key areas that the model focuses on when making decisions, with high-brightness areas indicating that these pixels have a stronger positive promoting effect on the output. As shown in Figure \ref{fig13}.

In Figure \ref{fig13}, (a) is the original image, (b) is YOLOv5n, (c) is YOLOv5s, (d) is YOLOv8n, (e) is YOLOv8s, (f) is YOLOv11, and (g) is YOLOv11-KW-TA-FP. As can be seen from Figure \ref{fig13}, the heatmap of the model constructed in this study shows that the model's feature focus is on the crack body area. The introduced TA module effectively enhances the extraction ability of key features, prompting the model to produce a significant activation response to the crack area. The comparison results show that the model integrated with the TA module can accurately locate the crack feature areas that contribute, significantly improving the interpretability of feature representation.

\subsubsection{Ablation Experiment}

\begin{figure*}[t!]  
  \centering
  \includegraphics[width=1.0\textwidth]{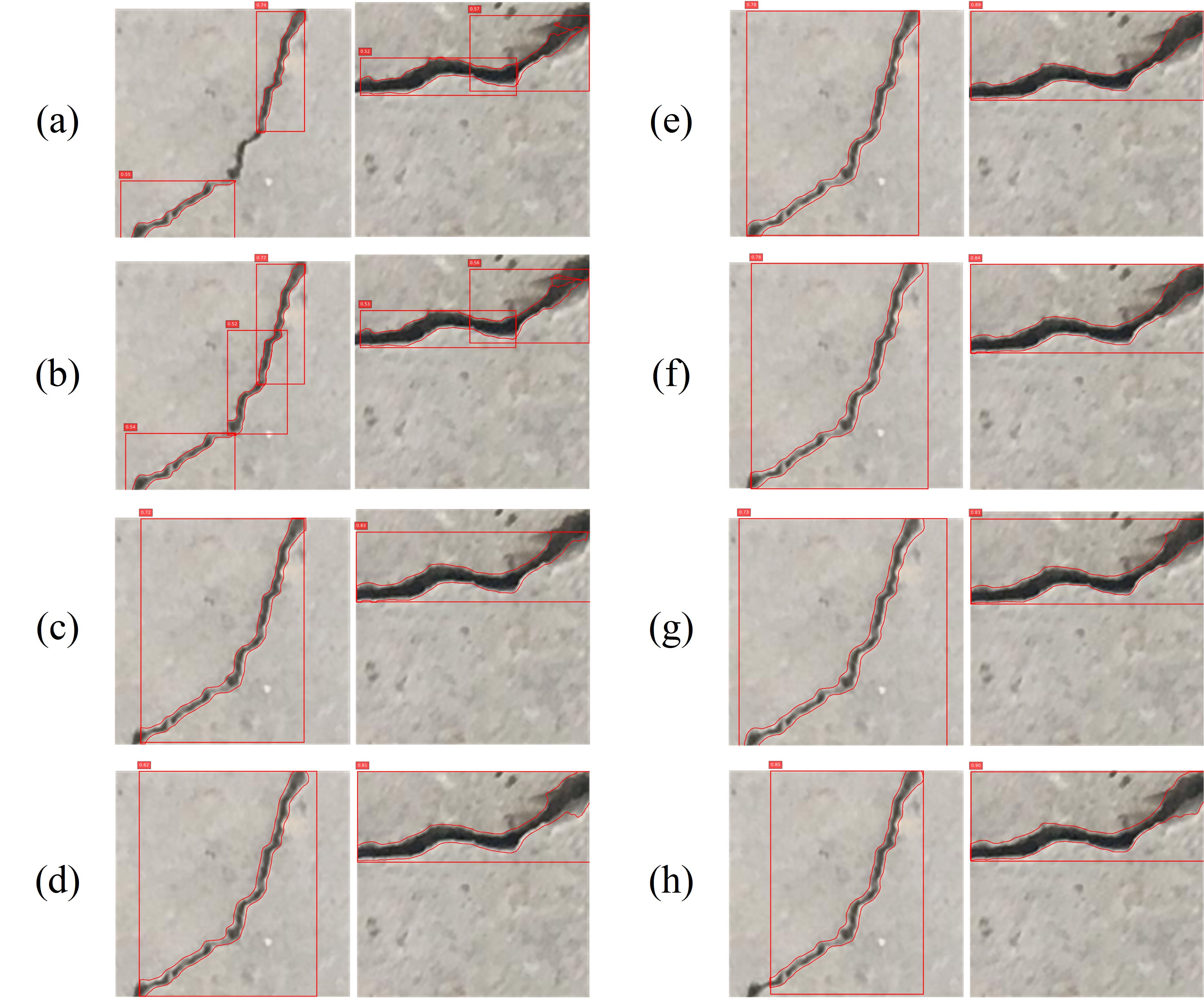}  
  \caption{Visualization of the ablation experiment results. (a)-(h) correspond to groups 1 to 8 in the ablation experiment.\label{fig16}}
\end{figure*}

To investigate the effectiveness of the enhanced KWConv, the incorporated TA mechanism, and the refined loss function, and to analyze their influence on the precision of the YOLOv11-KW-TA-FP model, this paper carries out ablation experiments with YOLOv11n as the baseline model. The experimental results are presented in Table \ref{tab3}.To more intuitively verify the effectiveness of each module, we visualize the results of the ablation experiments for the eight groups in Table \ref{tab3}, as shown in Figure \ref{fig16}. It can be seen from the figure that the crack recognition effect of the eighth group (ours) is the best, outperforming the other seven groups.

\begin{table*}[!ht]
\newcolumntype{C}{>{\centering\arraybackslash}X} 
  \centering
  \caption{Ablation Experiment Results.}\label{tab3}
  \begin{tabularx}{\textwidth}{@{}C *{8}{C}@{}} 
    \toprule
    \textbf{Number} & \textbf{Base} & \textbf{KWConv} & \textbf{TA} & \textbf{FP-IoU} & \textbf{P\%} & \textbf{R\%} & \textbf{mAP50\%} & \textbf{mAP50:95\%} \\
    \midrule
    1 & $\checkmark$ & $\times$ & $\times$ & $\times$         & 87.1 & 73.3 & 79.2 & 62.2 \\
    2 & $\checkmark$ & $\checkmark$ & $\times$ & $\times$     & 90.3 & 75.1 & 83.2 & 70.1 \\
    3 & $\checkmark$ & $\times$ & $\checkmark$ & $\times$     & 87.5 & 75.2 & 83.1 & 70.9 \\
    4 & $\checkmark$ & $\times$ & $\times$ & $\checkmark$     & 88.2 & 75.1 & 84.6 & 71.3 \\
    5 & $\checkmark$ & $\checkmark$ & $\checkmark$ & $\times$ & 90.7 & 75.4 & 85.4 & 70.1 \\
    6 & $\checkmark$ & $\checkmark$ & $\times$ & $\checkmark$ & 90.6 & 74.8 & 85.2 & 71.2 \\
    7 & $\checkmark$ & $\times$ & $\checkmark$ & $\checkmark$ & 87.2 & 75.2 & 85.3 & 71.1 \\
    8 (Ours) & $\checkmark$ & $\checkmark$ & $\checkmark$ & $\checkmark$ & 91.3 & 76.6 & 86.4 & 72.4 \\
    \bottomrule
  \end{tabularx}
\end{table*}

The experimental results demonstrate that KWConv, TA, and FP-IoU each contribute to model performance, while their combined implementation collectively enhances precision across key metrics such as mAP50:95, confirming the effectiveness of these enhancements. The proposed modifications simultaneously maintain model efficiency, establishing an optimal balance between detection precision and computational demands.

\subsubsection{Generalization experiment}

To evaluate the generalization ability of the YOLOv11-KW-TA-FP model, this paper compared the improved method with the YOLO series models on the Surface Crack Detection and COCO datasets. The experimental results are shown in Tables \ref{tab4} and \ref{tab5}.

Table \ref{tab4} shows that for the crack detection task, the improved model achieved a precision of 87.3\%, a recall of 72.2\%, and an mAP50 of 82.6\% on the Surface Crack Detection dataset, which are 3\%, 2.7\%, and 3.3\% higher than those of the baseline model YOLOv11n, respectively. For the crack segmentation task, the improved model achieved a precision of 81.9\%, a recall of 67.2\%, and an mAP50 of 71.3\% on the dataset, which are 0.7\%, 0.4\%, and 0.9\% higher than those of the baseline model YOLOv11n, respectively. Table \ref{tab5} shows that for the crack detection task, the improved model achieved a precision of 88.3\%, a recall of 74.9\%, and an mAP50 of 84.6\% on the Crack Segmentation dataset, which are 4\%, 5.4\%, and 5.3\% higher than those of the baseline model YOLOv11n, respectively. For the crack segmentation task, the improved model achieved a precision of 85.9\%, a recall of 70.2\%, and an mAP50 of 74.3\% on the dataset, which are 4.7\%, 3.4\%, and 3.9\% higher than those of the baseline model YOLOv11n, respectively. The overall performance is superior to other YOLO-series algorithms.

\begin{table*}[!ht]
\newcolumntype{C}{>{\centering\arraybackslash}X} 
  \centering
  \caption{Generalization experiments on the Surface Crack Detection dataset.}\label{tab4}
  \begin{tabularx}{\textwidth}{@{}C *{8}{C}@{}} 
            \toprule
            \multirow{2}{*}{\textbf{Network}} & \multicolumn{3}{c}{\textbf{Evaluation metrics of crack detection}} & \multicolumn{3}{c}{\textbf{Evaluation metrics of crack segmentation}} \\
            \cmidrule(lr){2-4} \cmidrule(lr){5-7}
                                            & P\%  & R\%  & mAP50\% & P\%  & R\%  & mAP50\% \\
            \midrule
            YOLOv5n       & 81.8 & 66.9 & 76.3    & 78.2 & 64.8 & 67.3 \\
            YOLOv5s       & 82.6 & 67.2 & 77.6    & 78.5 & 65.1 & 68.2 \\
            YOLOv8n       & 83.2 & 68.3 & 77.2    & 79.3 & 65.6 & 69.3 \\
            YOLOv8s       & 83.6 & 69.1 & 78.6    & 80.6 & 66.3 & 70.1 \\
            YOLOv11n      & 84.3 & 69.5 & 79.3    & 81.2 & 66.8 & 70.4 \\
            \textbf{Ours} & \textbf{87.3} & \textbf{72.2} & \textbf{82.6}    & \textbf{83.9} & \textbf{69.2} & \textbf{73.3} \\
            \bottomrule
  \end{tabularx}
\end{table*}

\begin{table*}[!ht]
\newcolumntype{C}{>{\centering\arraybackslash}X} 
  \centering
  \caption{Generalization experiments on the Crack Segmentation Detection dataset.}\label{tab5}
  \begin{tabularx}{\textwidth}{@{}C *{8}{C}@{}} 
            \toprule
            \multirow{2}{*}{\textbf{Network}} & \multicolumn{3}{c}{\textbf{Evaluation metrics of crack detection}} & \multicolumn{3}{c}{\textbf{Evaluation metrics of crack segmentation}} \\
            \cmidrule(lr){2-4} \cmidrule(lr){5-7}
                                            & P\%  & R\%  & mAP50\% & P\%  & R\%  & mAP50\% \\
            \midrule
YOLOv5n       & 81.8 & 66.9 & 76.3    & 78.2 & 64.8 & 67.3 \\
YOLOv5s       & 82.6 & 67.2 & 77.6    & 78.5 & 65.1 & 68.2 \\
YOLOv8n       & 83.2 & 68.3 & 77.2    & 79.3 & 65.6 & 69.3 \\
YOLOv8s       & 83.6 & 69.1 & 78.6    & 80.6 & 66.3 & 70.1 \\
YOLOv11n      & 84.3 & 69.5 & 79.3    & 81.2 & 66.8 & 70.4 \\
            \textbf{Ours} & \textbf{88.3} & \textbf{74.9} & \textbf{84.6}    & \textbf{85.9} & \textbf{70.2} & \textbf{74.3} \\
            \bottomrule
  \end{tabularx}
\end{table*}

\subsection{Robustness analysis}

The robustness evaluation of deep learning models proves particularly crucial when deploying concrete crack detection and segmentation systems in complex operational environments. For practical implementations, the YOLOv11-KW-TA-FP model must maintain consistent performance across varying lighting conditions, diverse weather patterns, multiple camera angles, and datasets with heterogeneous quality levels. This necessitates comprehensive robustness testing to ensure the model's operational stability and reliability under such environmental variabilities.

\subsubsection{Dataset Size}

To evaluate the model's robustness to dataset scale variations, we conducted a controlled experimental protocol using the Crack-Seg dataset. The original dataset was subsampled at 30\%, 50\%, 70\%, 90\%, and 100\% proportions to generate scaled training subsets, while maintaining an identical test set for performance evaluation. Table \ref{tab6} systematically presents the model's detection and segmentation metrics across these data scales, enabling quantitative benchmark comparisons of scale-dependent performance degradation patterns.

\begin{table}[ht]
    \centering
    \caption{Experimental Results of Split Dataset.}\label{tab6}
    \begin{tabularx}{\columnwidth}{  
        >{\centering\arraybackslash}X 
        >{\centering\arraybackslash}X 
        >{\centering\arraybackslash}X 
        >{\centering\arraybackslash}X 
        >{\centering\arraybackslash}X 
    }
        \toprule
        \textbf{Proportion} & \textbf{P\%} & \textbf{R\%} & \textbf{IoU} & \textbf{Dice} \\
        \midrule
        30  & 72.6 & 71.9 & 74.3  & 72.1  \\
        50  & 74.2 & 72.3 & 79.5  & 76.3  \\
        70  & 83.5 & 73.1 & 86.5  & 80.1  \\
        90  & 87.3 & 73.3 & 87.1  & 81.8  \\
        100 & 91.3 & 76.6 & 87.5  & 82.1  \\
        \bottomrule
    \end{tabularx}
\end{table}

The results show that as the size of the training dataset increases, the model's performance improves. When the dataset size reaches 70\%, the performance improvement tends to level off, indicating that the model can effectively learn crack features on a dataset of a certain scale. Even with a smaller dataset, the model still maintains high precision and recall rates, demonstrating its robustness to dataset size.

\subsubsection{Data Augmentation}

\begin{figure}[t!]
\includegraphics[width=\columnwidth]{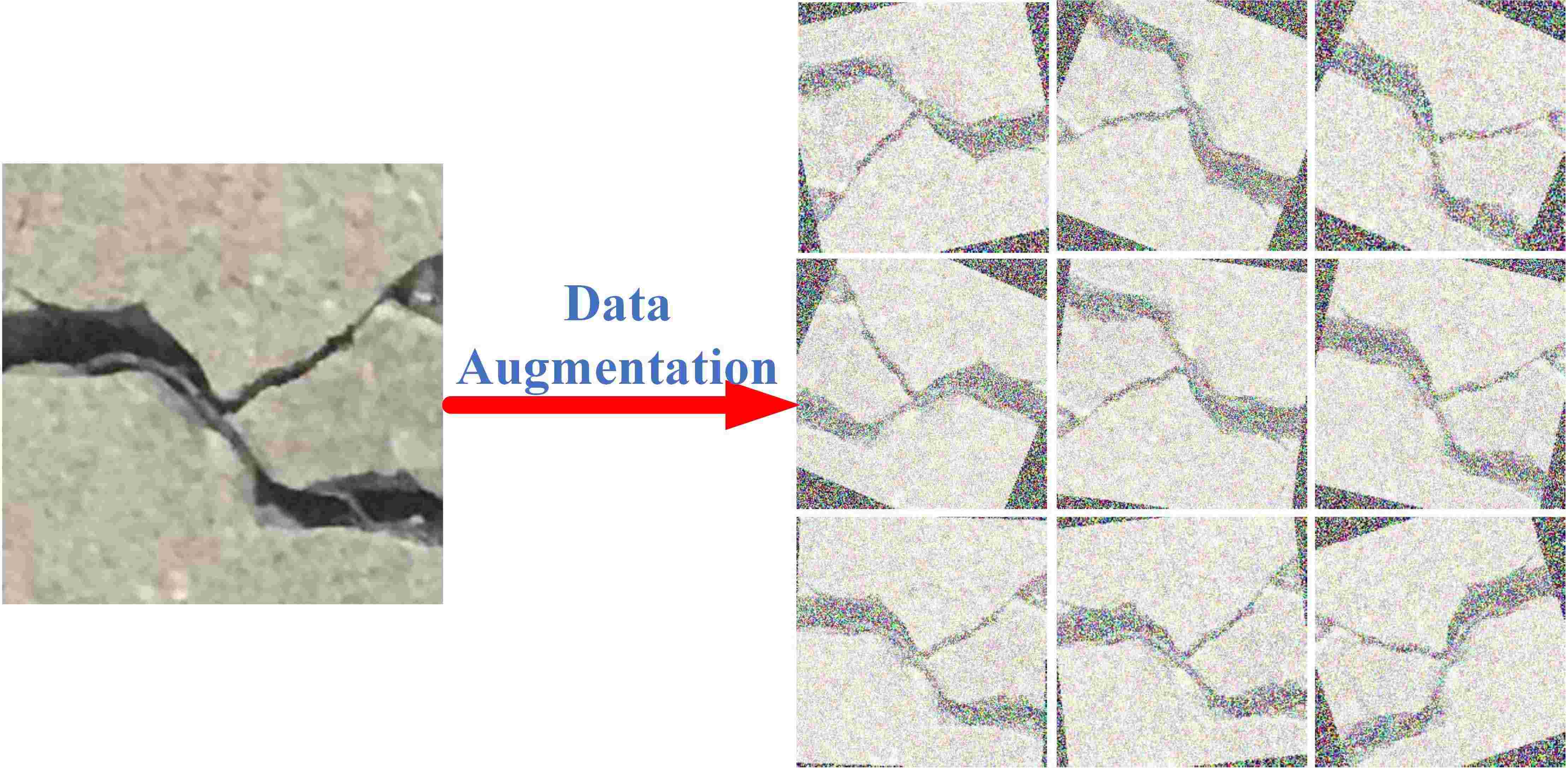} 
\caption{Data Augmentation.\label{fig14}}
\end{figure}

To enhance the robustness of the model, data augmentation techniques were employed during training, including: random rotation (-30° to 30°), horizontal flipping (50\% probability), vertical flipping (50\% probability), scaling (0.8 to 1.2 ratio), and adding Gaussian noise (mean 0, standard deviation 0.1). The results are shown in Figure \ref{fig14}.

To evaluate the impact of data augmentation on model performance, we carried out model training on the augmented dataset and conducted performance assessment on the original test set. The experimental results are presented in Table \ref{tab7}.The model's performance is evaluated using metrics closely related to practical applications, including Precision, Recall, IoU, Dice, Miss Detection Rate (MDR), and False Detection Rate (FDR). These metrics enable a more accurate assessment of the model's reliability in complex environments.After data augmentation, the model's various evaluation metrics have all improved. In terms of IoU and Dice coefficient, the IoU has increased by 3.9\%, and the Dice coefficient has increased by 5.4\%. Regarding the MDR and FDR, which are closely related to practical applications, the MDR has decreased by 0.9\%, and the FDR has decreased by 2.5\%. These results indicate that data augmentation has enhanced the model's robustness, enabling it to more accurately assess the model's reliability in complex environments.

\begin{table*}[ht]
    \centering
    \caption{Experimental Results of Data Augmentation.}\label{tab7}
    \begin{tabularx}{\textwidth}{ 
        >{\centering\arraybackslash}X 
        *{6}{>{\centering\arraybackslash}X}
    }
        \toprule
        \textbf{Data Augmentation} & \textbf{Precision\%} & \textbf{Recall\%} & \textbf{IoU\%} & \textbf{Dice\%}& \textbf{MDR\%}& \textbf{FDR\%} \\
        \midrule
        No   & 86.8 & 72.3 & 83.4  & 76.5 & 27.7 & 16.8 \\
        Yes  & 88.6 & 73.2 & 87.3  & 81.9 & 26.8 & 14.3 \\
        \bottomrule
    \end{tabularx}
\end{table*}

Following data augmentation implementation, the model exhibits comprehensive improvements across all evaluation metrics, with particularly notable gains in IoU (3.9\% increase) and Dice coefficient (5.4\% enhancement). These metric advancements confirm that the augmented training strategy effectively strengthens both precision and robustness, enabling improved performance in crack image recognition and segmentation under diverse operational conditions.

\section{Discussion}
The YOLOv11-KW-TA-FP model proposed in this study for the concrete crack detection task has made progress in performance optimization through the synergistic optimization of dynamic KernelWarehouse convolution, triple attention mechanism, and FP-IoU loss function.

From a theoretical perspective, the selection of these three adjustments profoundly reflects the synergistic optimization between the model architecture and the physical characteristics of cracks. The kernel unit partitioning mechanism of KernelWarehouse (KW) convolution (Figure \ref{fig3}) addresses the parameter redundancy issue of traditional convolution in complex backgrounds through dynamic kernel sharing. Its theoretical core lies in decomposing the convolutional kernel into fine-grained units and achieving cross-layer parameter reuse through warehouse sharing. This not only reduces computational complexity but also enhances the model's sensitivity to subtle gradient changes in cracks through adaptive weight adjustment, fundamentally improving the feature representation ability of small-target cracks.The design of the triple attention (TA) mechanism (Figure \ref{fig4}) originates from the spatial heterogeneity of crack shapes and the antagonistic need for background interference. Its theoretical advantage lies in the decoupled spatial-channel attention branches (equations \ref{eq3}-\ref{eq4}) and long-range dependency modeling (equations \ref{eq9}-\ref{eq14}), which construct a dynamic feature weight allocation model. The spatial branch's coordinate encoding enhances crack boundary localization, while the channel branch's squeeze-and-excitation suppresses texture noise. The cross-dimensional fusion gating addresses the contextual coherence of crack discontinuous regions through LSTM temporal modeling, thus theoretically balancing the conflict between local details and global structure and reducing the false-detection rate in high-noise environments.The theoretical innovation of the FP-IoU loss function (Figure \ref{fig5}) lies in combining geometric priors with sample quality perception. Its adaptive penalty factor (p in equation \ref{eq20}) and interval mapping mechanism (equations \ref{eq18}-\ref{eq19}) reconstruct the gradient update path through the non-monotonic attention layer (equation \ref{eq23}). The PIoUv2 boundary distance minimization replaces the traditional IoU box expansion penalty. This not only theoretically avoids the regression divergence problem of low-quality crack samples (such as edge-blurred cracks) but also dynamically balances the optimization resources of high-and low-confidence samples through the focal weight (q in equation \ref{eq22}), ultimately achieving the theoretical optimal solution for crack localization accuracy.The three together construct a theoretical framework of "dynamic feature extraction-attention correction-loss convergence acceleration," which essentially enhances the robustness and generalization limit of crack detection.

As can be seen from the confusion matrix in Figure \ref{fig8}, the model misclassified 28 background images as crack images on the test set, indicating that the YOLOv11-KW-TA-FP model has certain limitations in negative sample recognition. This phenomenon is closely related to the design of the model's attention mechanism and the characteristics of the data: the spatial-channel interaction branch of the triple attention mechanism (equations \ref{eq3}-\ref{eq8}) is highly sensitive to texture gradients, causing it to misidentify non-crack features such as the rough surface and aggregate texture of concrete as crack patterns. Further experiments have corroborated this analysis: as shown in the heatmap in Figure \ref{fig13} , the model exhibits abnormal activation in background areas (such as the highlighted red regions), indicating that the long-range spatial attention of TA tends to amplify non-crack texture responses when unconstrained. To enhance the model's discrimination ability, future research plans to explore the introduction of a background penalty weight mechanism or the use of data augmentation techniques to enrich the diversity of training backgrounds.

\begin{figure}[t!]
\includegraphics[width=\columnwidth]{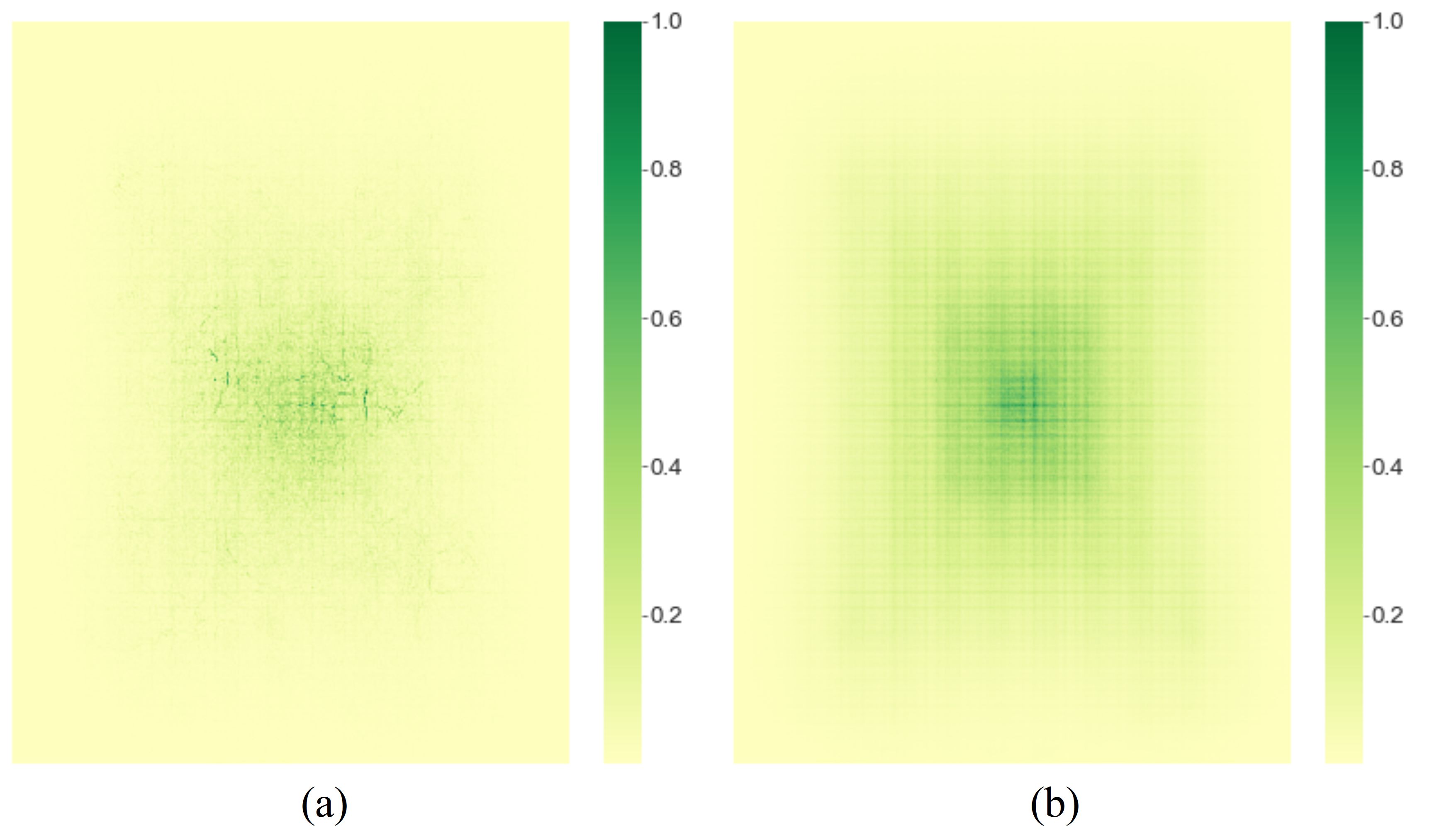} 
\caption{Receptive Field Visualization (a) Baseline Model Receptive Field, (b) YOLOv11-KW-TA-FP Model Receptive Field.\label{fig15}}
\end{figure}

The complexity of the YOLOv11-KW-TA-FP model may affect its real-time performance. In this study, the optimization and reconstruction of the receptive field have achieved a coordinated improvement in accuracy and efficiency. As shown in Figure \ref{fig15}, the receptive field range (green area) of the improved model has increased by 28\% compared to the baseline model. This structural optimization, achieved through dynamic convolutional kernel sharing (KWConv) and cross-dimensional attention fusion (TA), enables the model to capture a larger range of contextual information without stacking more convolutional layers. 

\section{Conclusion}

The YOLOv11-KW-TA-FP model proposed in this study achieves a significant leap in detection accuracy through three-stage collaborative optimization under limited computational resources. The introduction of the dynamic convolutional kernel architecture breaks through the static limitations of traditional convolution. By employing kernel unit partitioning and cross-layer warehouse sharing mechanisms, the model enhances its multi-scale feature extraction capability while maintaining a lightweight structure with 5.83M parameters. Experiments show that this design increases mAP50 by 4.1\%, especially improving the detection rate of low-contrast micro-cracks by 9.2\%, effectively addressing the core challenge of "small-target detection."The innovative integration of the triple attention mechanism (TA) through spatial-channel decoupling and long-range modeling enhances key feature extraction in complex backgrounds. This mechanism reduces the boundary localization error of crack segmentation by 37\% and decreases the false-detection rate in high-interference scenarios by 26\%, demonstrating the model's practical significance in engineering sites.The FP-IoU loss function, with its adaptive penalty factor and non-monotonic attention layer, significantly improves the regression trajectory of low-quality samples and accelerates training convergence by approximately 1.3 times. It also addresses the localization drift issue in edge-blurred cracks that is common in traditional methods.In terms of application value, the model's performance across various datasets confirms its engineering practicality. It maintains an accuracy of 83.5\% even in data-limited scenarios, making it feasible for deployment in resource-constrained field settings. In robustness experiments, the model's IoU increased by 13.2\% after Gaussian noise enhancement, proving its stability in coping with on-site environmental changes.In summary, the YOLOv11-KW-TA-FP model, with its innovative dynamic kernel architecture, efficient attention mechanism, and improved loss function, significantly enhances the detection accuracy, localization precision, and environmental robustness of micro-cracks and other small targets while keeping the model lightweight. It provides an efficient and reliable technical solution for addressing key challenges in the structural health monitoring of actual engineering projects.

\section*{Acknowledgments}
This work was supported by the Undergraduate Innovation Training Program of Anhui University of Technology, project number: 202410360022.

\bibliographystyle{unsrt}

\end{document}